\pdfoutput=1

\documentclass[11pt]{article}

\usepackage[]{emnlp2021}

\usepackage{times}
\usepackage{latexsym}

\usepackage[T1]{fontenc}

\usepackage[utf8]{inputenc}

\usepackage{microtype}

\usepackage[]{graphicx}	
\usepackage{amsmath} 
\usepackage{amsfonts} 
\usepackage{amssymb} 
\usepackage{amsthm} 
\usepackage{exscale} 
\usepackage{amstext} 
\usepackage{array} 
\usepackage{rotating} 
\usepackage[noend,linesnumbered,ruled,english]{algorithm2e} 
\usepackage{longtable} 

\usepackage{booktabs}
\usepackage{multirow}
\usepackage{hhline}

\DeclareMathOperator*{\argmax}{arg\,max}
\DeclareMathOperator*{\argmin}{arg\,min}

\def\bigtablesep{-0.05cm}

\title{Active Learning for Argument Mining: A Practical Approach}

\author{
    Nikolai Solmsdorf\Thanks{N.S. did this work as part of his master's thesis at CIS.},
    Dietrich Trautmann, 
    Hinrich Sch{\"u}tze\\
    Center for Information and Language Processing (CIS)\\
    Ludwig Maximilian University (LMU), Munich, Germany\\
    {\tt nikolai.solmsdorf@zos.badw.de; dietrich@trautmann.me}\\
}

\begin{document}

\maketitle

\begin{abstract}

Despite considerable recent progress, the creation of well-balanced and diverse resources remains a time-consuming and costly challenge in Argument Mining. Active Learning reduces the amount of data necessary for the training of machine learning models by querying the most informative samples for annotation and therefore is a promising method for resource creation. In a large scale comparison of several Active Learning methods, we show that Active Learning considerably decreases the effort necessary to get good deep learning performance on the task of Argument Unit Recognition and Classification (AURC).
\end{abstract}

\section{Introduction}

Despite the great many advancements the field of Argument Mining brought about, the task of retrieving and classifying argumentative segments in heterogene, \textit{open} text remains an intricate one. Among the persisting challenges, e.g., which argument model should be applied in the first place or the search for a common evaluation scheme, we find the need for quality data with expert annotation most salient. As in particular the creation of new training data is a costly endeavour, the motivation of the following paper is to lay out how the annotation and training procedure for such a specialised corpus from the field of Argument Mining, i.e., the AURC corpus \cite{Trautmann2020}, can be designed more efficiently. A sophisticated reduction of the amount of data necessary for the training by querying the most informative samples for annotation is a promising method to this end, i.e., a method commonly referred to as Active Learning (henceforth AL). 

\begin{figure}[htb]
	\includegraphics[width=0.5\textwidth]{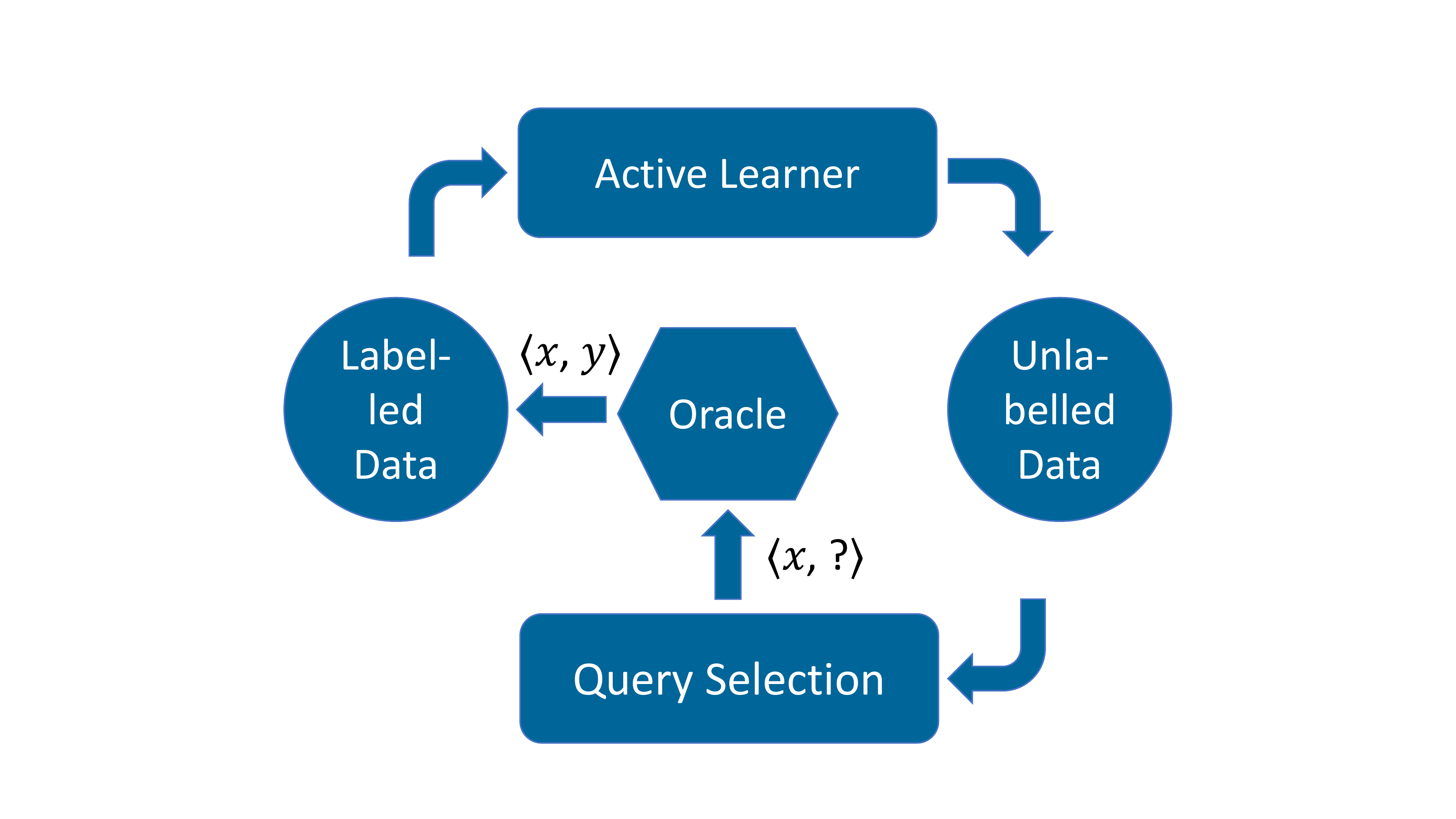}
	\caption{Active Learning Process}
	\label{fig:related:al_schema}
\end{figure}

Hence, the core of this paper is made up of a comprehensive comparison of various model architectures for the AURC task—i.e., recognising argumentative segments in text on the token level—together with a pragmatically motivated selection of AL strategies—from simple Uncertainty sampling to various clustering-based methods in order to cater for Diversity and Representativeness sampling to a novel transfer learning approach, boosted by a unique focus on a so-called warm-start scenario for an AL procedure. It proves successfully that a reduction of samples required for an optimal model performance is not only feasible but can lead to a markedly abridged training and annotation process, thus serving as an exemplary case study for the design of future corpora.

\section{Related Work}

\subsection{Active Learning}
Active Learning seeks to achieve better performance of an algorithm for machine learning with less
training, if it is allowed to select the data it learns from. Hence, through a \textit{query selection} process an \textit{active learner}, such as a model for sentence classification or for sequence labelling, tries to cleverly choose the most informative instances which are then labelled by an \textit{oracle}, e.g., a human annotator, so that the model can be trained as quickly and effectively as possible (Figure \ref{fig:related:al_schema}).

The degree of informativeness is evaluated in the query selection step of the AL loop, and the sampling strategy deployed during this step determines the criterion of informativeness, hence, functioning as an integral aspect of an AL scenario. Thus, varying query selection strategies received much research attention, and a plethora of policies have been proposed \cite{Settles2009,Settles2012,Yang2016,Munro2020}. They broadly address either one or a combination of the following query criteria: (i) Uncertainty or Confusion, i.e., where is the model most uncertain in its prediction, classically measured, e.g., through Least Confidence \cite{Lewis1994a,Lewis1994}, Margin Confidence \cite{Scheffer2001ActiveHM}, Entropy \cite{Settles2008}, ensemble methods, such as Query-by-Committee \cite{Seung1992}, decision-theoretic approaches, like Expected Error Reduction \cite{Roy2001}, or, in more recent Deep Learning approaches, by Monte Carlo Dropout \cite{Gal2017,Siddhant2018} or through Learning Loss \cite{Yoo2020}; (ii) Representativeness, i.e., when sampling a batch of instances in each query step—the so-called batch-mode scenario—redundancy has to be avoided and samples from each region of the input space should be queried, through, e.g., some density-weighted method, such as clustering \cite{Shen04,Settles2008,Huang2014}; and (iii) Diversity, i.e., an extension of the Representativeness approach, where “odd” instances from the input space are included as well \cite{Sener2017,Zhdanov2019,Munro2020}.
Additionally, as of recently, so-called agnostic methods that are less dependent on human heuristics or do not rely on such strategies at all have received considerable attention, such as a Multi-Armed Bandit approach, Active Learning by Learning \cite{Hsu2015}, or Data-driven AL through Reinforcement Learning \cite{Bachmann2017,Pang2018,Konyushkova2019}.
One commonality of these diverse strategies is: Each policy is virtually independent from the base model used as the learner, e.g., for classification—i.e., theoretically. In practice, however, and in particular during the liaison of Active with Deep Learning a serious bottleneck is encountered: As for each AL episode the learner has to be retrained, and the training of Deep Learning architectures usually is computationally heavy, joining the two approaches efficiently poses a challenge. A practical and pragmatic approach for a “hard” task, such as a structured prediction like sequence labelling \cite{akhundov2018sequence}, thus is warranted.

\subsection{Argument Mining}

It is apparent that the automatic detection and identification of argumentative propositions in text are very challenging objectives, especially in regard of being able to generalise across heterogenous texts which are not inherently argumentative \cite{Palau2009,Moens2013,Lippi2015,Ajjour2018}. A recent method thus moves from the closed-domain discourse level approach to Argument Mining, to a so-called \textit{information seeking} concept, where the aim is to identify self-contained argumentative propositions on the sentence level relevant to a specific \textit{topic} drawn from diverse sources, which are not inherently argumentative \cite{Stab2018}—nonetheless, current contributions still endeavour on classifying deeper argumentative structures \cite{Habernal2017,Kuribayashi2019,Miller2019}. A simple definition of an argument thus suffices: It is a span of text providing evidence or reasoning to the given topic, i.e., \textit{argument units}. This concept can be interpreted as a “flat” version of a discourse level approach: The topic of the text can be seen as an implicit claim or conclusion—in literature, sometimes the topic is also referred to as the \textit{major claim}—and all argumentative statements function either as support or opposition to the topic. Detecting argumentative statements on the \textit{token} rather than on the \textit{sentence level} was proposed to further advance this information seeking approach and to alleviate some of its limitations \cite{Trautmann2020}: Separate argumentative propositions are not merged into a single argument, so that a larger number of arguments can be retrieved; opposing stances in the argumentation in a single sentence can be distinguished; errors in sentence segmentation decrease; and finally, complex argumentative sentences are broken down into simpler argument units. Applying sequence labelling models—a CRF and BiLSTM were used, with the addition of various BERT models—hence brought considerable improvements about, in particular in cross-domain experiments. Further, this led to the creation of the so-called \textit{Argument Unit Recognition and Classification} (AURC) corpus.

\section{Data Set}

\subsection{The AURC Corpus} 
The premise for the AURC corpus was to produce a well-balanced collection of argumentative propositions, reflecting both, supporting and opposing stances, as well as non-argumentative statements drawn from freely available and heterogenous sources. Further, the statements should pertain to a selection of eight controversial topics: (i) \textit{abortion}, (ii) \textit{cloning}, (iii) \textit{marijuana legalisation}, (iv) \textit{minimum wage}, (v) \textit{nuclear energy}, (vi) \textit{death penalty}, (vii) \textit{gun control}, and (viii) \textit{school uniform}. This topic assortment was drawn from \cite{Stab2018}, and, by virtue of its polemical nature, is meant to ensure easily discernible argumentative statements. The corpus contains $1{,}000$ sentences per topic, i.e., in total $8{,}000$ instances, which were tapped from a \textit{Common Crawl} snapshot and indexed with \textit{Elasticsearch}. The time-consuming particularities which led to the creation of the corpus and its exact statistics can be tracked in \cite{Trautmann2020}.\footnote{The current version of the corpus can be downloaded from \url{https://github.com/trtm/AURC} (last accessed July 21, 2021).}

\subsection{Input Encodings}

A selection of the pre-trained versions of the most common word embeddings as well as current contextualised embedding models are used as encodings of the input from the AURC corpus, in order to evaluate their respective impact on the models employed for the sequence labelling task \cite{akhundov2018sequence,Lu2019}. This selection includes:

(i) For token representations: (a) \textit{Word2vec} –  \textit{Google News} corpus \cite{Mikolov2013,Mikolov2013a}; (b) \textit{GloVe} – \textit{Common Crawl} \cite{Pennington2014}; (c) \textit{FastText} – \textit{Common Crawl} \cite{Bojanowski2016,Joulin2017,Mikolov2017}; (d) BERT\textsubscript{BASE[CTX]} and BERT\textsubscript{LARGE[CTX]} – \textit{BookCorpus} \& English Wikipedia \cite{Devlin2019BERTPO}.

(ii) For sentence representations, which are part of various clustering experiments (outlined below): (a) For Word2vec, GloVe, and fastText, a simple fixed-length mean vector of all the word vectors in one sequence is computed; (b) BERT\textsubscript{BASE[CLS]} and BERT\textsubscript{LARGE[CLS]}; the special classification token ([CLS]) is the final hidden state vector and can be used as the aggregate sequence representation for classification tasks; (c) SBERT\textsubscript{BASE[M-NLI]} and SBERT\textsubscript{LARGE[M-NLI]}—refined sentence representations derived from BERT after  \cite{Reimers2019b}.

\section{Experimental Set-Up}

In order to evaluate various AL strategies for the AURC sequence labelling task in a practical setup—i.e., the AL loops need to run in an efficient computational environment\footnote{Cf. the Appendix for technical details, further information to the model setup, as well as the various strategies.}—, which allows state-of-the-art strategies and Deep Learning architectures to be considered as well, the experimental system consists of the following components:

\subsection{Active Learner}

This set of models comprises standard as well as more complex model architectures for the comparison in the AL setting. \textit{Nota bene}: As of late, \textit{fine-tuning} Transformer architectures \cite{Vaswani2017}, such as BERT, has brought about great success in many different NLP applications. However, the training and fine-tuning of such models is computationally heavy, in particular, for structured prediction tasks, so that, here it was opted to leave these architectures out—however, please note, that in less complex tasks, e.g., text classification, fine-tuned Transformer models were successfully employed in an AL setting, albeit with simple sampling strategies such as Uncertainty \cite{Schroeder2021}. Nonetheless, in the face of the constraints and in order to cater for the latest development still as much as possible, the contextualised input encodings which can be derived from BERT have been included; hence, these encodings were not fine-tuned but “frozen” for the pass through the respective model architectures. This led to the following choice of models:

(i) A combination of a BiLSTM with a \textit{Character Level CNN} as input encoding, and a final CRF layer for decoding, i.e., the \textit{BiLSTM-CNN-CRF} \cite{Ma2016}.

(ii) A simpler \textit{BiLSTM-CRF} \cite{Huang2015, Lample2016} which skips the Character Level CNN as the additional input encoding.

(iii) And a so-called \textit{LinCRF}, i.e., a combination of a single \textit{Linear Layer} with a CRF for decoding, as a less complex model which may provide a pragmatic solution to the task.

\subsection{Active Learning Policies}

The following set of sampling strategies comprises standard as well as more complex methods in order to be able to compare different approaches. The AL scenario studied in the experiments is the \textit{batch-mode} \cite{Hoi2006,WangZhengandYe2015} and \textit{pool-based} setting \cite{Lewis1994a}, while three general sampling criteria are covered, i.e., \textit{Uncertainty}, \textit{Diversity}, and \textit{Representativeness} (cf. above). In addition, the very beginning of an AL process is scrutinised:

(i) \textbf{Cold- vs. Warm-Start}: Commonly, for the compilation of the first set of instances which are labelled by the oracle before the AL process starts, samples are randomly chosen from the unlabelled dataset, i.e., a setting that is also referred to as \textit{Cold-Start}. Nonetheless, there have been strategies proposed for an alternative approach, i.e., the so-called \textit{Warm-Start}, which rely on clustering as a means for constructing an informative first set of instances which is able to \textit{kick-start} the AL process \cite{Kang2004,Nguyen2004,Hu2010,Zhu2008,Zhu2010,Munro2020}. Here, a similar approach is adopted; however, a comparison of various clustering algorithms—i.e., \textit{k-Means} \cite{Steinhaus1956,Lloyd1982}, \textit{DBSCAN} \cite{Ester1996}, and \textit{Agglomerative Hierarchical Clustering} \cite{Voorhees1986,Hamdouchi1989}—together with the selection of input encodings introduced above is also included, which in this form has not been investigated before. 

(ii) \textbf{Random Sampling}: At the query selection step of each AL episode, new samples are chosen at random from the unlabelled pool of data. This baseline method has been found to work surprisingly well \cite{Cawley2011,Yang2016}, and all other sampling strategies are measured against it.

(iii) \textbf{Uncertainty Sampling}: It is the most basic, well-studied, and also often exceptionally successful sampling strategy \cite{Yang2016}. In each loop, new unlabelled samples are chosen on the basis of one the following uncertainty criteria: (a) Least Confidence; (b) Margin Confidence; and (c) Entropy.

(iv) \textbf{Cluster Random Sampling}: Using clustering is a mean to include diverse samples from differing regions of the input space, thus, avoiding sampling redundant instances. At each query selection step, the unlabelled data is clustered with one of the aforementioned cluster algorithms, after which, new samples are chosen randomly and equally from each cluster in order to form a new batch for labelling by the oracle.

(v) \textbf{Cluster Uncertainty Sampling}: It aims at choosing those samples which come from diverse regions of the input space and are able to induce confusion to the model. Thus, in every AL episode, samples are clustered, and subsequently, each cluster is indexed according to one of the three uncertainty criteria introduced above; the most uncertain instances are sampled equally from each cluster to compose the new batch of unlabelled data.

(vi) \textbf{Cluster Representative Sampling}: It chooses instances which lie in or near the centre of each cluster, and thus \textit{represent} each cluster region best—\textit{nota bene}, there are varying definitions of \textit{representativeness}, e.g., \cite{Munro2020} measures the unlabelled data against the already labelled. The method introduced here uses the clusters calculated at the query selection step, determines the core regions of each cluster, and samples equally from each core region until the new batch of unlabelled instances is complete.

(vii) \textbf{Cluster Diversity Sampling}: This strategy, inversely, chooses not from the centre but the fringes of each cluster. Its aim is to examine if those samples that are difficult to group and classify by a clustering algorithm, e.g., noise points in the case of DBSCAN, are able to add meaning to an informative batch of new samples.

(viii) \textbf{ATLAS}: A complex method which specifically addresses AL in combination with Deep Learning is \textit{Active Transfer Learning} (ATL) \cite{Munro2020}. Its gist is to \textit{transfer} the output of the main classification model, i.e., the learner, to train a second model on the objective of predicting correctly versus incorrectly classified instances. Then, the second binary classification model is applied on the unlabelled data in order to predict which samples are most likely classified incorrectly. Those samples are eventually selected for the next AL episode by means of some standard uncertainty criterion. In summary, this is an elaborate, alternative approach to Uncertainty Sampling which combines the predictions of two Deep Learning models for the query selection step. \textit{Active Transfer Learning for Adaptive Sampling} (ATLAS) \cite{Munro2020} is a further advancement: A single sampling criterion—such as Uncertainty by means of ATL—might lead to a selection of instances which are too similar. ATLAS addresses this issue so that it \textit{adaptively} samples \textit{diverse} instances within one query selection step. The basic setup is as follows: 1) The learner is trained on a subset of the available data; 2) the trained model is used for an evaluation on a validation data set; 3) the results of the learner are transferred as it is captured which validation instances were classified correctly and incorrectly; 4) the newly classified validation data becomes the new training set for a second binary classification model with the objective to predict correct and incorrect labels; 5) the newly trained second binary model is used to classify the remaining unlabelled data; 6) a fraction of the targeted total set for an AL batch is selected—e.g., if the total size of the batch of samples from the query selection step is $100$, ATLAS only choses $10$ (both parameters, i.e., the total batch size and the ATLAS fraction have to be set manually). Next, ATLAS follows a simple assumption: Those instances which are chosen in the query selection will be labelled by the oracle and thus become part of the training set, the main model is re-trained on this data, and thus, those newly added instances presumably would be labelled correct. Hence, 7) the labels of the $10$ samples that have been chosen based on the binary classification as incorrect are set to correct and added for the time being to the training set for the binary classification model, i.e., the validation set of the main model; 8) still within the overarching query selection step, the binary model is re-trained on the validation set that has been increased with the 10 samples from ATLAS step (6), i.e., ideally, the binary classification model now has learned more; 9) the re-trained binary model is used to predict another $10$ incorrect samples from the unlabelled data. The steps 6) to 9) are repeated until the targeted batch size, i.e., in this example, $100$, is reached. Then, 10) the total set of queried instances is passed on to the oracle, which labels the data, and the next AL episode starts. The main idea here is, that through the incremental re-training of the binary classification model within one AL loop, samples are selected that are dissimilar to each other and thus convey the diversity criterion on top of uncertainty. The original proposal of ATLAS \cite{Munro2020} suggests, in the query selection step, to choose the incorrectly labelled instances by means of a standard uncertainty criterion, i.e., Least or Margin Confidence, or Entropy. However, in the following experiments, at each ATLAS step (6), the samples are selected randomly from the incorrectly labelled data, as it is assumed that a further induction of diversity can be achieved through this; therefore, this implementation of ATLAS represents a slight deviation.\footnote{Cf. the Appendix for the ATLAS algorithm.}

To the best of our knowledge, this study thus is the first of its kind, as former research did touch on AL with sequence labelling, however, either the comparison of the model architectures is lacking, cf. e.g., \cite{Settles2008,Deng2018,Wang2018,Munro2020,Liu2020}, or the testing of differing sampling strategies is limited \cite{Erdmann2019}.

\subsection{Metrics}
In order to evaluate the performance of the sequence labelling models on detecting and identifying argumentative statements on the token level, the same metric is adopted as in the original AURC paper, i.e., the \textit{Macro} $F_1$ measure averaged across sequences.

In order to evaluate the performance of each sampling strategy, it is observed how many training samples are needed to reach the following thresholds: $90\%$, $95\%$, $99\%$, and $100\%$ of the \textit{Macro} $F_1$ baseline results obtained through the training with the full training set (in-domain: $4{,}200$, cross-domain: $4{,}000$ samples). The fewer training instances required to reach these thresholds, the better the AL strategy.

All experiments are conducted  in- and cross-domain on the AURC data, using the original dataset splits, with this hyperparameter setting: (i) The experiments run over $10$ different random seeds for non-deterministic methods \cite{Reimers2017}; (ii) the mean of all 10 runs is reported; (iii) \textit{mini-batch training} \cite{Masters2018} is used for the Deep Learning architectures, with the batch size set to $64$; (iv) the maximum number of training epochs for the neural network models is set to $100$; the training epochs of the binary model within ATLAS are set to $10$; (v) \textit{early stopping} \cite{Caruana2000} is used for the neural models, observing the system's main metric, i.e., the \textit{Macro} $F_1$, with a minimum number of training epochs of $10$, and the patience before early stopping breaks the training loop is also set to $10$ epochs.

\begin{table}[!h]\centering
\scriptsize
\resizebox{0.49\textwidth}{!}{
\begin{tabular}{|l||c|c|}
    \hhline{|-||-|-|}
    \begin{tabular}{lrl}
         &  & \multicolumn{1}{r}{} \\
        \midrule
         &  & \multicolumn{1}{l}{model} \\
        \midrule
        \multirow{3}{*}{\rotatebox[origin=c]{90}{in}}
         & 3 & BiLSTM-CNN-CRF\textsubscript{AL} \\[\bigtablesep]
         & 4 & BiLSTM-CRF\textsubscript{AL} \\[\bigtablesep]
         & 5 & LinCRF\textsubscript{AL} \\
        \midrule
        \multirow{3}{*}{\rotatebox[origin=c]{90}{cross\!}}
         & 6 & BiLSTM-CNN-CRF\textsubscript{AL} \\[\bigtablesep]
         & 7 & BiLSTM-CRF\textsubscript{AL} \\[\bigtablesep]
         & 8 & LinCRF\textsubscript{AL} \\
    \end{tabular}&
    \begin{tabular}{c}
        \multicolumn{1}{r}{} \\
        \midrule
        dev \\
        \midrule
        .677 \\[\bigtablesep]
        .676 \\[\bigtablesep]
        .662 \\[\bigtablesep]
        \midrule
        .538 \\[\bigtablesep]
	.523 \\[\bigtablesep]
        .503 \\
    \end{tabular}&
    \begin{tabular}{c}
       	\multicolumn{1}{r}{} \\
        \midrule
        test \\
        \midrule
        .647 \\[\bigtablesep]
        .656 \\[\bigtablesep]
        .615 \\
        \midrule
        .525 \\[\bigtablesep]
        .499 \\[\bigtablesep]
        .428 \\
    \end{tabular}\\
    \hhline{|-||-|-|}
\end{tabular}
}
\vspace{2mm}
\caption{Summarised results, token $F_1$ on AURC dev and test for AL sequence labelling models}
\label{tab:al_model_baselines}
\end{table}

\section{Results \& Discussion}

Through a series of experiments the following findings could be established:

(i) Five variants for encoding the input on the word level were compared for each model architecture (cf. above), showing—unsurprisingly— that the contextualised word embeddings derived from BERT models outperform standard embeddings, such as GloVe.\footnote{Cf. Table \ref{tab:model_baselines} in the Appendix.} This comparison led to optimal model/input encoding combinations for both AURC data splits, i.e., the BiLSTM-CNN-CRF with BERT\textsubscript{LARGE[CTX]}, the BiLSTM-CRF with BERT\textsubscript{LARGE[CTX]}, and the LinCRF with BERT\textsubscript{BASE[CTX]}, dubbed BiLSTM-CNN-CRF\textsubscript{AL}, BiLSTM-CRF\textsubscript{AL}, and LinCRF\textsubscript{AL}, respectively (Table \ref{tab:al_model_baselines}).

\begin{table}[hb]\centering
\scriptsize
\resizebox{.49\textwidth}{!}{
\begin{tabular}{|l||c|c|}
    \hhline{|-||-|-|}
    \begin{tabular}{lrlr|}
         &  & \multicolumn{1}{r}{} \\
        \midrule
         &  & \multicolumn{1}{l}{model} \\
        \midrule
        \multirow{3}{*}{\rotatebox[origin=c]{90}{in}}
         & 3 & BiLSTM-CNN-CRF\textsubscript{AL} \\[\bigtablesep]
         & 4 & BiLSTM-CRF\textsubscript{AL} \\[\bigtablesep]
         & 5 & LinCRF\textsubscript{AL} \\
        \midrule
        \multirow{3}{*}{\rotatebox[origin=c]{90}{cross\!}}
         & 6 & BiLSTM-CNN-CRF\textsubscript{AL} \\[\bigtablesep]
         & 7 & BiLSTM-CRF\textsubscript{AL} \\[\bigtablesep]
         & 8 & LinCRF\textsubscript{AL} \\
    \end{tabular}&
    \begin{tabular}{c}
        \multicolumn{1}{c}{Cold-Start} \\
        \midrule
        dev \\
        \midrule
        .483 \\[\bigtablesep]
        .450 \\[\bigtablesep]
        .358 \\
        \midrule
        .390 \\[\bigtablesep]
        .364 \\[\bigtablesep]
        .295 \\
    \end{tabular}&
     \begin{tabular}{l}
        \multicolumn{1}{c}{Warm-Start} \\
        \midrule
        \multicolumn{1}{c}{dev} \\
        \midrule
        .491 (DBSCAN \& Word2vec) \\[\bigtablesep]
        .489 (KM \& Word2vec) \\[\bigtablesep]
        .466 (AC \& BERT\textsubscript{LARGE[CLS]}) \\
        \midrule
        .420 (DBSCAN \& fastText) \\[\bigtablesep]
        .409 (KM \& SBERT\textsubscript{BASE[M-NLI]}) \\[\bigtablesep]
        .397 (KM \& fastText) \\
    \end{tabular}\\
    \hhline{|-||-|-|}
\end{tabular}
}
\vspace{2mm}
\caption{Token $F_1$ for cold-start sampling batch ($n=64$) with Random Sampling and best warm-start sampling batch with Cluster Random Sampling}
\label{tab:cold_warm_start}
\end{table}

(ii) The Cold- vs. Warm-Start experiments in this paper, proved that through a preliminary clustering step, the performance of the learner at the start can be increased considerably. In particular, it was shown that through representing the unlabelled instances by means of sentence embeddings, and applying an optimised clustering algorithm to these representations, Random Sampling for the starting batch was virtually always outmatched. Furthermore, there is not one encoding/algorithm combination which bests all other combinations, and the choice of the input encoding and the clustering algorithm is dependent on the model used as the learner as well as on the data split. All in all, it was demonstrated that such a warm-start approach is a valuable addition to the AL procedure.

\begin{table*}[!h]\centering
\scriptsize
\resizebox{.9\textwidth}{!}{
\begin{tabular}{|l||c|c|c|c|}
    \hhline{|-||-|-|-|-|}
    \begin{tabular}{lrlr|}
         &  & \multicolumn{1}{r}{} \\
        \midrule
         &  & \multicolumn{1}{l}{model \& strategy} \\
        \midrule
        \multirow{3}{*}{\rotatebox[origin=c]{90}{in}}
         & 3 & BiLSTM-CNN-CRF\textsubscript{AL} \& ATLAS \\[\bigtablesep]
         & 4 & BiLSTM-CRF\textsubscript{AL} \& Uncertainty – Least Conf. \\[\bigtablesep]
         & 5 & LinCRF\textsubscript{AL} \& Uncertainty – Least Conf. \\
        \midrule
        \multirow{3}{*}{\rotatebox[origin=c]{90}{cross\!}}
         & 6 & BiLSTM-CNN-CRF\textsubscript{AL} \& ATLAS \\[\bigtablesep]
         & 7 & BiLSTM-CRF\textsubscript{AL} \& Cluster Uncertainty – Entropy \\[\bigtablesep]
         & 8 & LinCRF\textsubscript{AL} \& Cluster Representative \\
    \end{tabular}&
    \begin{tabular}{c}
        \multicolumn{1}{c}{90\%} \\
        \midrule
        dev \\
        \midrule
        320 \\[\bigtablesep]
        1024 \\[\bigtablesep]
        1408 \\
        \midrule
        256 \\[\bigtablesep]
        320 \\[\bigtablesep]
        384 \\
    \end{tabular}&
    \begin{tabular}{c}
        \multicolumn{1}{c}{95\%} \\
        \midrule
        dev \\
        \midrule
        640 \\[\bigtablesep]
        1920 \\[\bigtablesep]
        1984 \\
        \midrule
        640 \\[\bigtablesep]
        512 \\[\bigtablesep]
        768 \\
    \end{tabular}&
    \begin{tabular}{c}
        \multicolumn{1}{c}{99\%} \\
        \midrule
        dev \\
        \midrule
        2112 \\[\bigtablesep]
        2880 \\[\bigtablesep]
        2752 \\
        \midrule
        1280 \\[\bigtablesep]
        832 \\[\bigtablesep]
        1536 \\
    \end{tabular}&
     \begin{tabular}{l}
        \multicolumn{1}{c}{100\%} \\
        \midrule
        \multicolumn{1}{c}{dev} \\
        \midrule
        2432 \\[\bigtablesep]
        3392 \\[\bigtablesep]
        3328 \\
        \midrule
        1536 \\[\bigtablesep]
        1088 \\[\bigtablesep]
        1600 \\
    \end{tabular}\\
    \hhline{|-||-|-|-|-|}
\end{tabular}
}
\vspace{2mm}
\caption{Sampling thresholds (dev, $n$ samples) for AL sequence labelling models with best respective sampling strategy}
\label{tab:thresholds_best_strats}
\end{table*}

(iii) The performance of each and every strategy thus was evaluated for each sequence labelling model and AURC dataset split.\footnote{Cf. the Appendix for plots of the most successful strategies for each model.} Individually for each model, measured against the $100\%$ threshold, various strategies achieved the optimal model performance with the fewest samples, suggesting that the model architecture is a decisive factor for the choice of the sampling strategy (Table \ref{tab:thresholds_best_strats}). However, globally, i.e., measured at each threshold, in-domain, ATLAS outperforms the other policies at almost each and every threshold; cross-domain, in turn, the picture is more diverse, with ATLAS, again, as the strategy of choice for the most complex AL sequence labelling model, BiLSTM-CNN-CRF\textsubscript{AL}, and various strategies that which include clustering as the component for the other two models (Table \ref{tab:best_strats_thresholds_a}).

\begin{figure}[h]\centering
\begin{minipage}{.45\textwidth}\centering
	\includegraphics[width=1\textwidth]{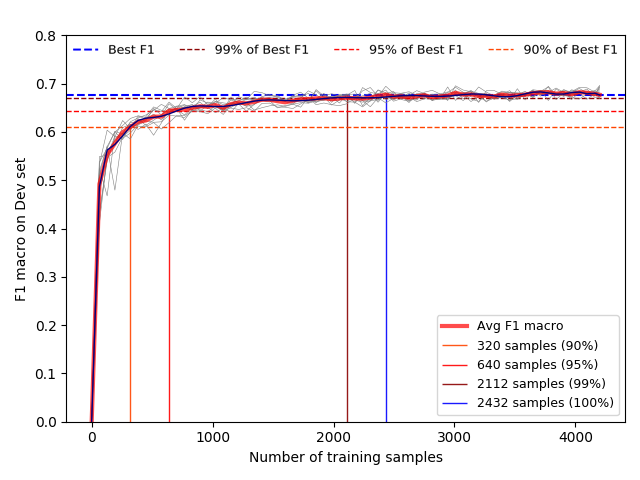}
	\caption{ATLAS with BiLSTM-CNN-CRF\textsubscript{AL} (dev, in-domain)}
	\label{fig:system:lcc_atlas_in}
\end{minipage}\hfill
\begin{minipage}{.45\textwidth}\centering
	\includegraphics[width=1\textwidth]{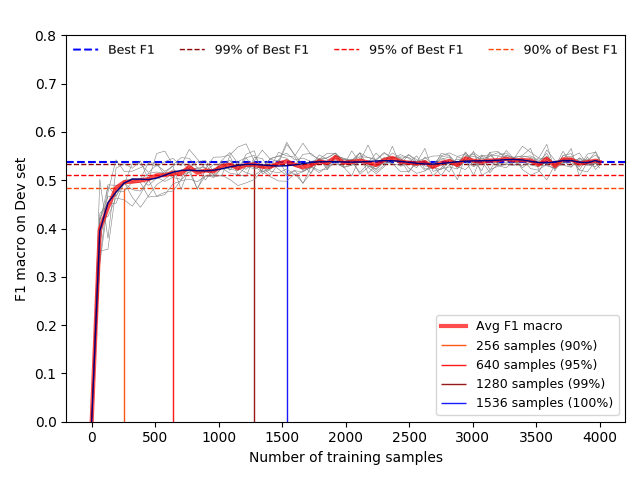}
	\caption{ATLAS with BiLSTM-CNN-CRF\textsubscript{AL} (dev, cross-domain)}
	\label{fig:system:lcc_atlas_cross}
\end{minipage}
\end{figure}

(iv) Reducing the amount of training samples is arguably an essential aspect of AL. However, AL aims at making the training and annotation process for a machine learning model comprehensively more efficient. The reduction of the time required for the training of the learner can be another valuable component for such a system, especially, if the learner comes from the field of Deep Learning, where training times typically are longer than in simpler machine learning architectures. Accordingly, this paper also examined the efficiency of the three AL sequence labelling models—all of which are Deep Learning architectures—in terms of training times. Surprisingly, the most light-weight model, i.e., the LinCRF\textsubscript{AL}, which was included in the experiments in particular for this comparison, all in all, did not fare well. After the BiLSTM-CNN-CRF\textsubscript{AL}, which, due to its complexity, needs the most training time, the LinCRF\textsubscript{AL} occupies a mediocre second place. The trade-off in accuracy that has to be taken into account with that model further disqualifies the architecture. The most surprising observation is, though, that the second best model, i.e., the BiLSTM-CRF\textsubscript{AL}, is considerably faster than any of the two other models, making it a strong recommendation as a reasonable compromise with an almost negligible performance drop-off in-domain compared to the BiLSTM-CNN-CRF\textsubscript{AL} and only slightly worse performance cross-domain. However, if the overall objective is to achieve the maximum in terms of accuracy with the AL system, the combination of the BiLSTM-CNN-CRF\textsubscript{AL} with ATLAS still has the advantage, in particular, taking into consideration that ATLAS manages to arrive at a near optimal performance earlier than the BiLSTM-CRF\textsubscript{AL}.

\begin{table}[!h]\centering
\scriptsize
\resizebox{0.49\textwidth}{!}{
\begin{tabular}{|l||c|}
    \hhline{|-||-|}
    \begin{tabular}{lrl}
         &  & \multicolumn{1}{r}{} \\
        \midrule
         &  & \multicolumn{1}{l}{model} \\
        \midrule
        \multirow{3}{*}{\rotatebox[origin=c]{90}{in}}
         & 3 & BiLSTM-CNN-CRF\textsubscript{AL} \\[\bigtablesep]
         & 4 & BiLSTM-CRF\textsubscript{AL} \\[\bigtablesep]
         & 5 & LinCRF\textsubscript{AL} \\
        \midrule
        \multirow{3}{*}{\rotatebox[origin=c]{90}{cross\!}}
         & 6 & BiLSTM-CNN-CRF\textsubscript{AL} \\[\bigtablesep]
         & 7 & BiLSTM-CRF\textsubscript{AL} \\[\bigtablesep]
         & 8 & LinCRF\textsubscript{AL} \\
    \end{tabular}&
    \begin{tabular}{c}
        \multicolumn{1}{r}{} \\
        \midrule
        epoch time \\
        \midrule
        20.04 \\[\bigtablesep]
        8.82 \\[\bigtablesep]
        6.84 \\[\bigtablesep]
        \midrule
        23.46 \\[\bigtablesep]
	9.12 \\[\bigtablesep]
        10.68 \\
    \end{tabular}\\
    \hhline{|-||-|}
\end{tabular}
}
\vspace{2mm}
\caption{Average training time per epoch (\textit{sec}) on full AURC dev for AL sequence labelling models}
\label{tab:al_model_avgtimes}
\end{table}

(v) Finally, a close look into the samples chosen in the query selection for each model/strategy combination and dataset split revealed further characteristics of the respective AL procedures and models:\footnote{Cf. the Appendix for data analysis plots.}

\begin{table*}[!h]\centering
\scriptsize
\resizebox{1\textwidth}{!}{
\begin{tabular}{|l||c|c|c|c|}
    \hhline{|-||-|-|-|-|}
    \begin{tabular}{lrlr|}
         &  & \multicolumn{1}{r}{} \\
        \midrule
         &  & \multicolumn{1}{l}{model} \\
        \midrule
        \multirow{3}{*}{\rotatebox[origin=c]{90}{in}}
         & 3 & BiLSTM-CNN-CRF\textsubscript{AL} \\[\bigtablesep]
         & 4 & BiLSTM-CRF\textsubscript{AL} \\[\bigtablesep]
         & 5 & LinCRF\textsubscript{AL} \\
        \midrule
        \multirow{3}{*}{\rotatebox[origin=c]{90}{cross\!}}
         & 6 & BiLSTM-CNN-CRF\textsubscript{AL} \\[\bigtablesep]
         & 7 & BiLSTM-CRF\textsubscript{AL} \\[\bigtablesep]
         & 8 & LinCRF\textsubscript{AL} \\
    \end{tabular}&
    \begin{tabular}{l}
        \multicolumn{1}{c}{90\%} \\
        \midrule
        \multicolumn{1}{c}{best strategy} \\
        \midrule
        ATLAS (320) \\[\bigtablesep]
        ATLAS (384) \\[\bigtablesep]
        ATLAS (640) \\
        \midrule
        Uncertainty – Entropy (192) \\[\bigtablesep]
        Uncertainty – Entropy (192) \\[\bigtablesep]
        Cluster Representative (384) \\
    \end{tabular}&
    \begin{tabular}{l}
        \multicolumn{1}{c}{95\%} \\
        \midrule
        \multicolumn{1}{c}{best strategy} \\
        \midrule
        ATLAS (640) \\[\bigtablesep]
        ATLAS (960) \\[\bigtablesep]
        ATLAS (1408) \\
        \midrule
        ATLAS (640) \\[\bigtablesep]
        Cluster Representative (448) \\[\bigtablesep]
        Cluster Representative (768) \\
    \end{tabular}&
    \begin{tabular}{l}
        \multicolumn{1}{c}{99\%} \\
        \midrule
        \multicolumn{1}{c}{best strategy} \\
        \midrule
        ATLAS (2112) \\[\bigtablesep]
        ATLAS (2816) \\[\bigtablesep]
        ATLAS (2752) \\
        \midrule
        ATLAS (1280) \\[\bigtablesep]
        Cluster Uncertainty –  Entropy (832) \\[\bigtablesep]
        Cluster Representative (1536) \\
    \end{tabular}&
     \begin{tabular}{l}
        \multicolumn{1}{c}{100\%} \\
        \midrule
        \multicolumn{1}{c}{best strategy} \\
        \midrule
        \textbf{ATLAS} (2432) \\[\bigtablesep]
        \textbf{Uncertainty – Least Conf.} (3392) \\[\bigtablesep]
        \textbf{Uncertainty – Least Conf.} (3328) \\
        \midrule
        \textbf{ATLAS} (1536) \\[\bigtablesep]
        \textbf{Cluster Uncertainty –  Entropy} (1088) \\[\bigtablesep]
        \textbf{Cluster Representative} (1600) \\
    \end{tabular}\\
    \hhline{|-||-|-|-|-|}
\end{tabular}
}
\vspace{2mm}
\caption{Best sampling strategy (with $n$ samples) for AL sequence labelling models for respective sampling thresholds}
\label{tab:best_strats_thresholds_a}
\end{table*}

(a) In-domain, about two thirds of the available unlabelled data are required to reach an optimal performance of the BiLSTM-CRF\textsubscript{AL} and the LinCRF\textsubscript{AL} with their respective AL sampling strategies, and the majority of the necessary training instances consists of non-argumentative segments, which suggests that even small portions of argumentative propositions can be enough for these models. The distribution of the polemical topics mostly reflects the token label distribution—this holds true for all model/strategy combinations across both AURC data splits. With respect to the average sentence length of the training instances, longer sentences apparently are more conducive to the training procedure. The BiLSTM-CNN-CRF\textsubscript{AL} with ATLAS, in turn, needs much less data—i.e., $\approx 18\%$—, and also the distribution of argumentative and non-argumentative propositions is quite different, implying that the few instances required for the efficient training should be largely argumentative. Furthermore, shorter samples suffice for this approach. (b) Cross-domain, across all three models, comparably, very few annotated samples are necessary to arrive at a near optimal or even optimal performance—i.e., $\approx 2-14\%$. For the BiLSTM-CNN-CRF\textsubscript{AL} model with ATLAS those samples, again, should be mostly argumentative; similarly, the LinCRF\textsubscript{AL} profits from argumentative propositions in the training set, though, at a smaller rate than the former model. The BiLSTM-CRF\textsubscript{AL}, in turn, generally needs relatively few argumentative segments. With regard to the average length of the sentences in the respective training sets, for the BiLSTM-CNN-CRF\textsubscript{AL}, short sentences suffice for an elevated performance, while the other two models benefit from longer sentences, as is the case in-domain.
 
In summary, and all things considered, i.e., the best input representation, a warm-start model, a reasonable model/strategy combination which also takes into account the time required for the training, and the characteristics of the instances queried, the following system for AL with sequence labelling is warrantable: (a) The contextualised text representations from BERT\textsubscript{LARGE} largely outmatched the other word embedding variants; (b) for the warm-start, no clear choice of sentence embedding and clustering algorithm suggests itself; yet, for its simplicity in use and also in the optimisation step, k-Means is recommended; for the sentence representation, various possibilities come into consideration, however, as it constitutes the state-of-the-art in this regard, Sentence Transformer representations have the edge; (c) one AL sampling strategy figures predominantly across both AURC data split, i.e., ATLAS; this uniquely designed query selection strategy also possesses some particularities in terms of the characteristics of its sampled instances—e.g., it favours short but argumentative sentences—and hence is recommended as the robust choice over the other strategies; (d) as described above, AL is all about efficiency; therefore, as it comes with the least amount of computational weight, the BiLSTM-CRF\textsubscript{AL} model architecture is recommended over the other two models.

\section{Future Work}
In addition to our findings, we identify the following promising avenues for future work on AL and sequence labelling, viz., (i) a study of the efficacy of the above-introduced system across heterogenous sequence labelling corpora, (ii) an even more cost-sensitive approach to AL taking into account real-world constraints, such as a \textit{noisy oracle} \cite{Donmez2008,Gupta2019}, \textit{budget and time}, together with annotation interfaces, annotator payment, and satisfaction \cite{Culotta2005,Munro2020}, and the \textit{stopping criterion} for an AL process cf. \cite{Laws2008,Zhang2018,Wu2019}, or (iii) agnostic Meta AL approaches in the context of Deep Learning, in particular, by means of Transformer architectures.
Furthermore, we will extend this work for aspect-based argument mining \cite{trautmann2020aspect}, relational approaches \cite{trautmann2020relational} and looking into a combination with domain adaptation techniques \cite{marz2019domain}.
\section{Conclusion}

To date, the composition and annotation of specialised corpora of open and heterogene text for Argument Mining remains a costly and time-consuming key challenge. This elaborate case study successfully showed on the basis of the AURC corpus and in the context of sequence labelling how through a comprehensive and practical Active Learning setup this challenge can be designed more efficiently, abridging the annotation process markedly and effectively. This includes a sophisticated warm-start setup at the beginning of an AL scenario, which makes use of state-of-the-art sentence embeddings and clustering algorithms to kick-start the process, an informed and pragmatic choice for the learner, i.e., a BiLSTM-CRF, and a recent query selection strategy which adaptively samples uncertain instances through transfer learning, i.e., ATLAS. This exemplary case study thus serves to alleviate the design of future corpora in the field of Argument Mining.


\section*{Acknowledgments}
We gratefully acknowledge support by Deutsche Forschungsgemeinschaft (DFG) (SPP-1999 Robust Argumentation Machines (RATIO), SCHU2246/13).

\cleardoublepage
\bibliography{anthology,custom}
\bibliographystyle{acl_natbib}
\appendix

\section{Appendix}
\label{sec:appendix}
\textbf{Technical Details} \\[0.2em]
Technically, all of the data preprocessing, the models and experiments are created with Python 3.7. The neural network models are implemented from scratch with \textit{PyTorch}, with the exception of the CRF layer, which is based on the \textit{pytorch-crf} library.\footnote{\url{https://pytorch-crf.readthedocs.io/en/stable/} (last accessed July 21, 2021)} The cluster algorithms, in turn, are based on the implementation of the scikit-learn library \cite{Pedregosa2011}.\footnote{\url{https://sklearn-crfsuite.readthedocs.io/en/latest/\#} (last accessed July 21, 2021)}  whereas the various Active Learning sampling strategies are likewise built in Python from the ground up.

The training of the models and all other experiments run on a single workstation with the following configuration: (i) CPU (Cores/Threads): i7-8700 @ 3.2GHz (6/12); (ii) Ram: 64GB; (iii) GPU: GeForce GTX 1060/6GB. PyTorch allows for efficient training on discrete GPUs, i.e., all the above-mentioned PyTorch models utilise the GPU for training. However, all model architectures were also successfully tested on a standard CPU (i7-4770HQ @ 2.2GHz [4/8]; RAM: 16GB), though, with the exception of the resource-heavy Active Learning experiments.

\medskip

\noindent \textbf{Models} \\[0.2em]
In the following, an overview of the models' detailed architecture and parameters is offered.

\begin{enumerate}

\item[(i)] \textit{BiLSTM-CNN-CRF}

This model comprises (a) an initial character embedding layer, with the embedding dimension set to $25$; (b) a character CNN module, which receives the character embeddings as input, applies a dropout layer ($0.3$) to it, and further a CNN with the \textit{window size} set to $3$ and the number of \textit{filters} to $30$, and a max pooling output layer; (c) a word embedding layer; (d) a BiLSTM module, i.e., two bidirectional LSTM layers, with the dimension of each (oneway) hidden LSTM layer set to $200$, and a \textit{dropout} rate of $0.5$ \cite{Hinton2012,Srivastava2014}; the concatenation of the outputs of the character CNN module and the word embedding layer—a dropout layer ($0.3$) is applied to both outputs before—functions as input for the BiLSTM module; and, finally, (e) a CRF layer which receives the output scores from the final linear layer of the BiLSTM module. For training, the loss function is defined by the CRF module, i.e., the \textit{negative log-likelihood} (NLL), whereas \textit{Adam} \cite{Kingma2015,Reddi2018} is selected as the optimiser for the models' parameters with a \textit{learning rate} of $0.001$.

\item[(ii)] \textit{BiLSTM-CRF}

Here, the CNN module is left aside, i.e., no character representation is included; thus, only word embeddings are used to encode the input. Other than that, the architecture is the same as in the BiLSTM module, with an additional CRF layer for decoding, and subsequently, NLL as the loss function for training. The optimiser remains unchanged.

\item[(vi)] \textit{LinCRF}

This “lightweight” model is included as an efficient alternative to the more complex architectures. It consists of a (a) word embedding layer, after which (b) a dropout layer ($0.3$) is applied, and a (c) simple linear layer; a final (d) CRF layer is added again for decoding. For training, again NLL is used as the loss function, whereas Adam is the optimiser with the learning rate set to $0.001$.

\end{enumerate}

\noindent \textbf{Sampling Strategies} \\[0.2em]
Further details on the sampling strategies are offered in the following section.
\begin{enumerate}
\item[(i)] \textit{Cold- vs. Warm-Start}

For the composition of the cold-start sampling batch, instances are chosen randomly from the unlabelled training set. For the warm-start scenario, in turn, three clustering algorithms are utilised, i.e., k-Means (KM), DBSCAN, and Agglomerative Hierarchical Clustering (AC). The following procedure describes the measures taken to this end:

\begin{enumerate}
\item[(a)] In order to cluster the unlabelled instances from the training set, each training instance is represented by a \textit{sentence embedding}. Accordingly, from each of the selected input encodings, sentence representations are derived: (1) For standard word embedding representations, i.e., Word2vec, GloVe, and fastText, a simple fixed-length mean vector of all the word vectors in one sequence is computed; (2) for BERT models, the [CLS] vector is used, i.e., BERT\textsubscript{BASE[CLS]} and BERT\textsubscript{LARGE[CLS]}; (3) furthermore, the representations created with Sentence Transformer are included as well, i.e., SBERT\textsubscript{BASE[M-NLI]} and SBERT\textsubscript{LARGE[M-NLI]}. All in all, the input for the clustering algorithms is represented with seven sentence representations for each data split, i.e., in- and cross-domain.

\item[(b)] High-dimensional representations can lead to poor and slow clustering performances; a dimensionality reduction algorithm, e.g., \textit{Principal Component Analysis} (PCA) is able to alleviate this problem \cite{Ding2004}.\footnote{\url{https://scikit-learn.org/stable/modules/clustering.html} (last accessed July 21, 2021)} \textit{Uniform Manifold Approximation and Projection} (UMAP) \cite{McInnes2018} has proven its superior performance over other dimension reduction techniques, and is adopted here as well. Thus, each sentence representation is reduced with UMAP to $2$ dimensions before the subsequent processing with the clustering algorithms.

\item[(c)] The three clustering algorithms are kept at their sklearn default settings, except for the following parameters: (1) KM: The \textit{k-Means++} variant is chosen, as it has shown better performance in the initialisation step of the cluster centres \cite{Celebi2013}; the randomisation of the initialisation is fixed, in order to enforce reproducibility; $k$ is optimised (cf. step [d]); (2) DBSCAN: The default value $5$ for the $minPts$ parameter is retained; $\varepsilon$ is optimised; (3) AC: No changes to the default; the distance criterion (\textit{distance threshold}) is optimised.

\item[(d)] The aforementioned hyperparameters are optimised in the following way: (1) Initial experiments with KM and DBSCAN predetermined a reasonable range in the hyperparameter space of $k$ and $\varepsilon$; for AC, in turn, after initial experiments, a dendrogram was plotted, in order to specify a range of values for the \textit{distance threshold} criterion, cf. Figure \ref{fig:system:ac_in_cls_l_dendro} for an example; the vertical axis represents the distance or dissimilarity between clusters, whereas the horizontal axis shows the objects and clusters; “great jumps” in the distance axis occurring before the fusion of two branches, can indicate meaningful cluster partitions; here, a common range of distances is chosen from the various dendrogram plots in order determine the optimal stopping criterion for the hierarchical clustering, i.e., the distance threshold which marks the most meaningful cluster partition; (2) The preliminary step resulted in a range for $k$ from $2$ to $16$ (stride 1), for $\varepsilon$ from $0.1$ to $0.51$ (stride $0.01$), and for the distance threshold from $10$ to $151$ (stride 5); (3) A combination of three scoring metrics is put into practice in order to evaluate the performance of the clustering algorithms across their respective hyperparameter space, i.e., the \textit{Silhouette Coefficient} \cite{Rousseeuw1987}, the \textit{Calinski-Harabasz Index} \cite{Calinski1974}, and the \textit{Davies-Bouldin Index} \cite{Davies1979}.

\begin{figure}[!htb]\centering
	\includegraphics[width=0.49\textwidth]{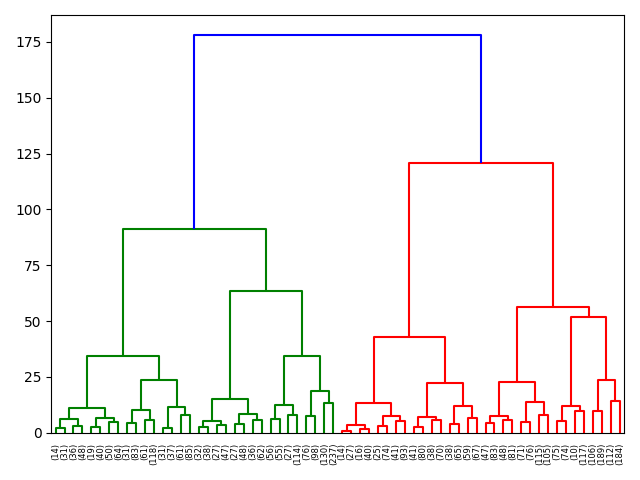}
	\caption[{Dendrogram of AC clustering of AURC data (in-domain) with \protect\linebreak BERT\textsubscript{LARGE[CLS]}}]{Dendrogram of AC clustering of AURC data (in-domain) with BERT\textsubscript{LARGE[CLS]}}
	\label{fig:system:ac_in_cls_l_dendro}
\end{figure}

(4) Based on the combination of the three metrics, the optimal values for the respective hyperparameters are determined through iterating 20 times over the parameter space of each cluster algorithm for each input encoding; the mean of the three resulting parameter values computed from each scoring metric establishes the final optimal value.

\item[(e)] Once the optimised clustering parameter values for each algorithm and input encoding are established, a warm-start batch of training instances is constructed. To this end, samples are chosen equally from each cluster region at random until the warm-start batch is complete.

\item[(f)] The newly established warm-start batch gets labelled, and is used for training with the sequence labelling models chosen for the Active Learning experiments, and the training process is evaluated with the whole development set. This is repeated for all possible combinations of input encodings and clustering algorithms, determining the optimal combination for a warm-start batch for each classification model.

\end{enumerate}

\item[(ii)] \textit{Random Sampling}

This constitutes the baseline for all of the following Active Learning sampling strategies. At each query selection step, instances are chosen at random from the whole unlabelled dataset until the sampling batch is complete. The optimal warm-start setting is not included; instead a cold-start is assumed, i.e., the first batch is also composed of randomly selected instances.

\item[(iii)] \textit{Uncertainty Sampling}

In order to compute the uncertainty, the output scores of the models chosen for the experiments are used. The scores are normalised with a \textit{softmax} function  \cite{Munro2020}; subsequently, the uncertainty is determined based on the following formulae:

(a) Least Confidence

\begin{equation} \label{eq:related:eq1}
\begin{split}
    x_{LC}^{*} & = \argmin\limits_{x_{i} \in\ \mathcal{U}} p_{\theta}(\hat{y} \mid x_{i})\\
    & = \argmax\limits_{x_{i} \in\ \mathcal{U}} 1 - p_{\theta}(\hat{y} \mid x_{i})
\end{split}
\end{equation}

(b) Margin Confidence

\begin{equation} \label{eq:related:eq2}
\begin{split}
    x_{MC}^{*} & = \argmin\limits_{x_{i} \in\ \mathcal{U}} \left[ p_{\theta}(\hat{y}_{1} \mid x_{i}) - p_{\theta}(\hat{y}_{2} \mid x_{i}) \right]\\
    & = \argmax\limits_{x_{i} \in\ \mathcal{U}} \left[ p_{\theta}(\hat{y}_{2} \mid x_{i}) - p_{\theta}(\hat{y}_{1} \mid x_{i}) \right]
\end{split}
\end{equation}

(c) Entropy

\begin{equation} \label{eq:related:eq3}
\begin{split}
    x_{H}^{*} & = \argmax\limits_{x_{i} \in\ \mathcal{U}} H_{\theta}(Y \mid x_{i})\\
    & = \argmax\limits_{x_{i} \in\ \mathcal{U}} -\sum\limits_{y\in\mathcal{Y}} p_{\theta}(y \mid x_{i}) \log p_{\theta}(y \mid x_{i})
\end{split}
\end{equation}

In order to capture the amount of uncertainty for a whole sequence based on the predictions for each token, the measures of uncertainty for each token prediction, which are computed with one of the three aforementioned criteria, are added together, and normalised through the respective sequence length; exemplified with entropy as the criterion for uncertainty, this amounts to the following, the so-called \textit{token entropy} (TE) \cite{Settles2008}:

\begin{equation} \label{eq:related:eq3}
\begin{split}
    \phi^{TE}(x) & = -\frac{1}{T} \sum\limits_{t=1}^{T} \sum\limits_{y\in\mathcal{Y}} p_{\theta}(y \mid x_{i}) \log p_{\theta}(y \mid x_{i})
\end{split}
\end{equation}

Hence, at each query selection step, the uncertainty is computed for all unlabelled instances. With the query batch size $n$, the $n$ most uncertain samples are selected for the composition of a new sampling batch.

\item[(iv)] \textit{Cluster Random Sampling}

This sampling strategy functions as an extension of the warm-start method: The optimised clustering algorithms from the warm-start setting are retained, and reused at each query selection step in order to sample instances equally from each cluster region at random until the sampling batch is complete.

\item[(v)] \textit{Cluster Uncertainty Sampling}

Here, the clustering algorithms from the warm-start experiment are reused also. However, at each query selection step, the cluster regions are further filtered with one of the three uncertainty criteria—the uncertainty within each cluster is computed as in the general Uncertainty Sampling strategy. Hence, from each region, equally, the most uncertain samples are selected until the sampling batch is complete.

\item[(vi)] \textit{Cluster Representative Sampling}

Again, the optimised cluster algorithms from the warm-start scenario are used in order to sample representative instances from each cluster region. To this end, at each query selection step the samples lying near or at the centre of each region are chosen equally from each cluster until the batch is complete. If k-Means is given as the optimal cluster algorithm, the central points can be determined easily by means of the sklearn implementation. In order to determine the centre regions of clusters derived from DBSCAN or AC, k-Means clustering is applied subsequently to each cluster ($k=1$).

\item[(vii)] \textit{Cluster Diversity Sampling}

This strategy is the inverse of the aforementioned method. At each query selection, the warm-start clustering algorithms are reused in order to sample outliers from each cluster region equally until the sampling batch is complete. Outlier detection is straightforward based on k-Means clustering; the instances which lie farthest from the centre are selected. Again, if DBSCAN or AC provide the optimal clustering algorithm, k-Means is applied afterwards to each cluster region ($k=1$).

\newpage

\item[(viii)] \textit{ATLAS}

The integral part at the query selection step with ATLAS is the transfer of the results of the main model to a secondary, binary classification model on the sentence level, which is used for the adaptive selection of divers and uncertain samples. Hence, in order to apply the second model, the results of the main model need to be bucketed in correct and incorrect predictions: The number of correct predictions for each sequence are summed up and normalised by the sequence length; if the ratio of correct predictions exceeds $0.5$, the sequence is bucketed as correct and vice versa. Thus, the binary model is trained on the newly classified sequences in order to distinguish correct from incorrect instances on the sentence level—for models with a BiLSTM module, the concatenation of the last hidden states from the final LSTM layer (forward and backward) functions as the sentence representation; for the LinCRF model, the sentence embedding representation derived from the original input encoding is used. The binary model, in turn, is a simple linear model consisting of two linear layers, with the hidden dimension size set to $128$. For its training, the \textit{binary cross entropy} with a \textit{sigmoid} activation is selected as the loss function (\textit{binary cross entropy with logits loss}), whereas Adam is chosen as the optimiser with a learning rate of $0.001$. At each adaptive query selection step, based on the prediction of the trained binary model, $8$ instances are drawn from the pool of unlabelled data at random; the adaptive step is repeated until the sampling batch is complete.

\end{enumerate}

\onecolumn

\begin{table}[!h]\centering
\scriptsize
\resizebox{1\textwidth}{!}{
\begin{tabular}{|l||c|c|c|c|c|}
    \hhline{|-||-|-|-|-|-|}
    \begin{tabular}{lrlr|r|r|}
         &  & \multicolumn{1}{r}{} \\
        \midrule
         &  & \multicolumn{1}{l}{model} \\
        \midrule
        \multirow{3}{*}{\rotatebox[origin=c]{90}{in}}
         & 3 & BiLSTM-CNN-CRF \\[\bigtablesep]
         & 4 & BiLSTM-CRF \\[\bigtablesep]
         & 5 & LinCRF \\
        \midrule
        \multirow{3}{*}{\rotatebox[origin=c]{90}{cross}}
         & 6 & BiLSTM-CNN-CRF \\[\bigtablesep]
         & 7 & BiLSTM-CRF \\[\bigtablesep]
         & 8 & LinCRF \\
    \end{tabular}&
    \begin{tabular}{c}
        \multicolumn{1}{c}{Word2vec} \\
        \midrule
        dev \\
        \midrule
        .583 \\[\bigtablesep]
        \textbf{.615} \\[\bigtablesep]
        .535 \\
        \midrule
        .448 \\[\bigtablesep]
        \textbf{.459} \\[\bigtablesep]
        .379 \\
    \end{tabular}&
    \begin{tabular}{c}
        \multicolumn{1}{c}{GloVe} \\
        \midrule
        dev \\
        \midrule
        .617 \\[\bigtablesep]
        \textbf{.631} \\[\bigtablesep]
        .538 \\
        \midrule
        \textbf{.460} \\[\bigtablesep]
        .459 \\[\bigtablesep]
        .388 \\
    \end{tabular}&
    \begin{tabular}{c}
        \multicolumn{1}{c}{fastText} \\
        \midrule
        dev \\
        \midrule
        .601 \\[\bigtablesep]
        \textbf{.626} \\[\bigtablesep]
        .536 \\
        \midrule
        .444 \\[\bigtablesep]
        \textbf{.448} \\[\bigtablesep]
        .386 \\
    \end{tabular}&
    \begin{tabular}{c}
        \multicolumn{1}{c}{BERT\textsubscript{BASE[CTX]}} \\
        \midrule
        dev \\
        \midrule
        \textbf{.665} \\[\bigtablesep]
        \textbf{.665} \\[\bigtablesep]
        .662 \\
        \midrule
        \textbf{.527} \\[\bigtablesep]
        .512 \\[\bigtablesep]
        .503 \\
    \end{tabular}&
    \begin{tabular}{c}
        \multicolumn{1}{c}{BERT\textsubscript{LARGE[CTX]}} \\
        \midrule
        dev \\
        \midrule
        \color{cyan}\textbf{.677} \\[\bigtablesep]
        .676 \\[\bigtablesep]
        .642 \\
        \midrule
        \color{cyan}\textbf{.538} \\[\bigtablesep]
        .523 \\[\bigtablesep]
        .482 \\
    \end{tabular}\\
    \hhline{|-||-|-|-|-|-|}
\end{tabular}
}
\vspace{2mm}
\caption{Token $F_1$ on AURC dev. Bold: Best performance per column and split (in-domain, cross-domain). Blue: Best overall performance.}
\label{tab:model_baselines}
\end{table}

\begin{figure}[!h]\centering
\begin{minipage}{.45\textwidth}\centering
	\includegraphics[width=1\textwidth]{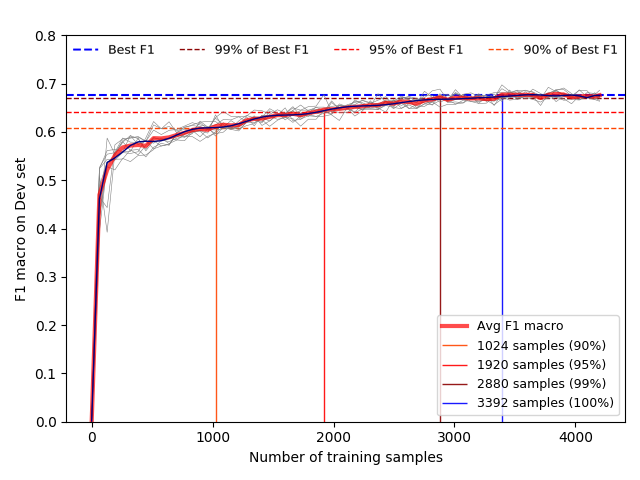}
	\caption{Uncertainty Sampling (Least Confidence) with BiLSTM-CRF\textsubscript{AL} (dev, in-domain)}	
	\label{fig:system:lc_cl_lc_in}
\end{minipage}\hfill
\begin{minipage}{.45\textwidth}\centering
	\includegraphics[width=1\textwidth]{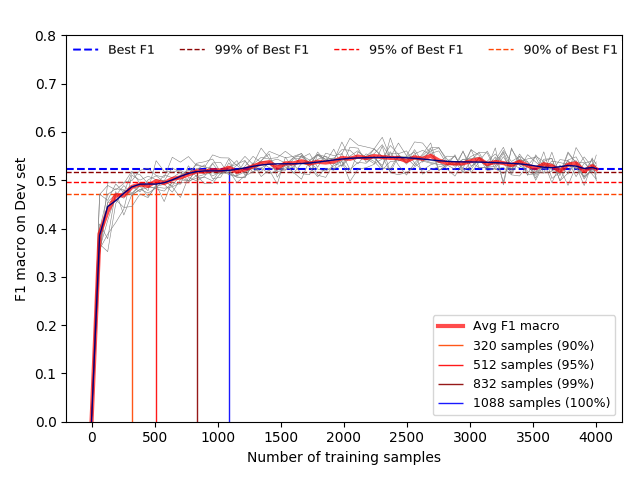}
	\caption{Cluster Uncertainty Sampling (Entropy) with BiLSTM-CRF\textsubscript{AL} (dev, cross-domain)}
	\label{fig:system:lc_cl_ent_cross}
\end{minipage}
\end{figure}

\begin{figure}[!h]\centering
\begin{minipage}{.45\textwidth}\centering
	\includegraphics[width=1\textwidth]{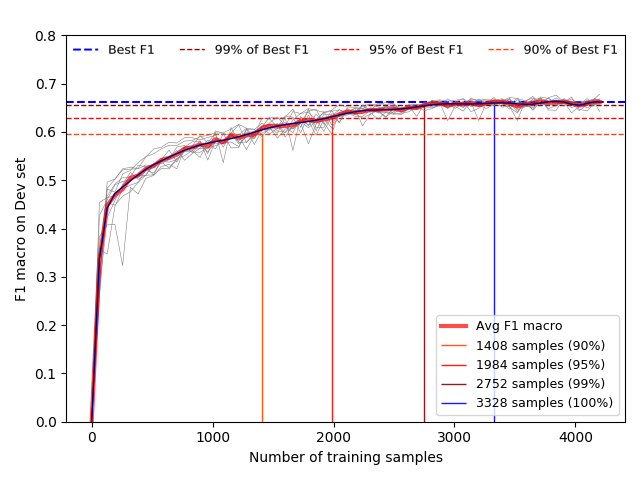}
	\caption[{Uncertainty Sampling (Least Confidence) with LinCRF\textsubscript{AL} (dev, in- \protect\linebreak domain)}]{Uncertainty Sampling (Least Confidence) with LinCRF\textsubscript{AL} (dev, in-domain)}
	\label{fig:system:lin_lc_in}
\end{minipage}\hfill
\begin{minipage}{.45\textwidth}\centering
	\includegraphics[width=1\textwidth]{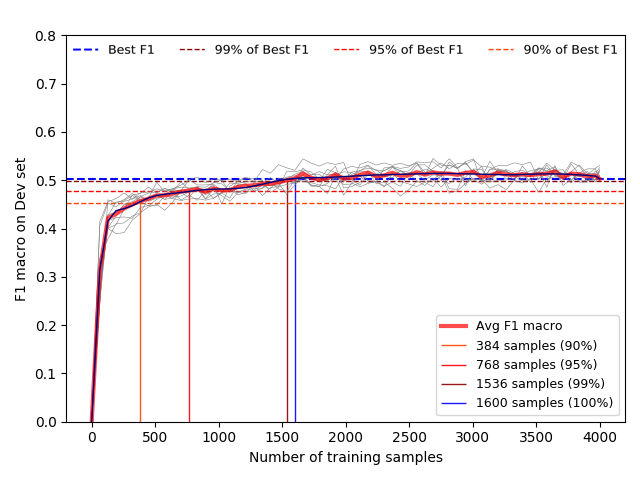}
	\caption{Cluster Representative Sampling with LinCRF\textsubscript{AL} (dev, cross-domain)}
	\label{fig:system:lin_cl_rpr_cross}
\end{minipage}
\end{figure}

\begin{figure}[!h]\centering
\begin{minipage}{.45\textwidth}\centering
	\includegraphics[width=1\textwidth]{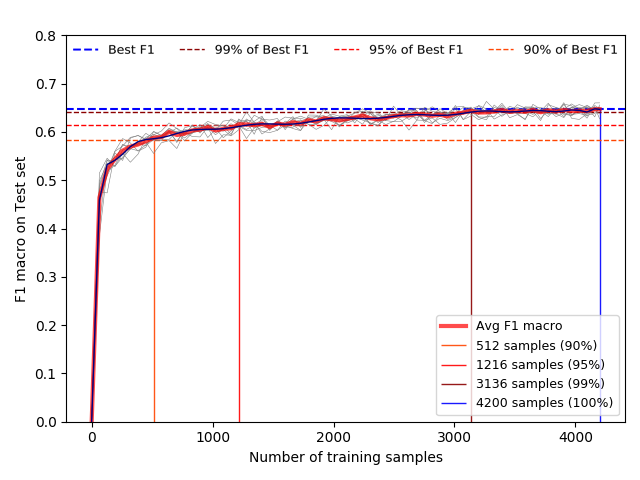}
	\caption{ATLAS with BiLSTM-CNN-CRF\textsubscript{AL} (test, in-domain)}
	\label{fig:system:lcc_atlas_in_test}
\end{minipage}\hfill
\begin{minipage}{.45\textwidth}\centering
	\includegraphics[width=1\textwidth]{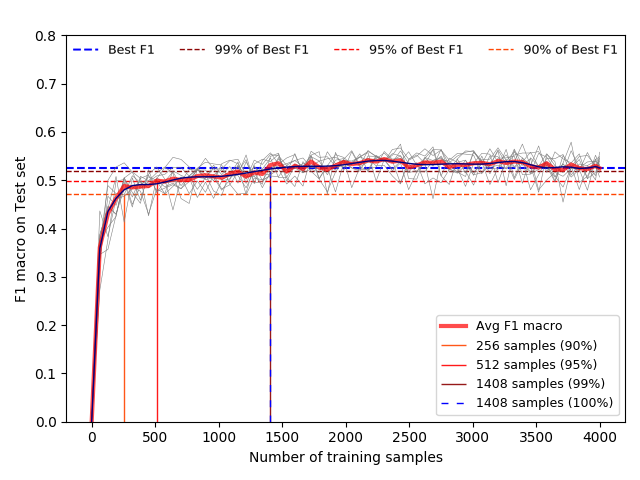}
	\caption{ATLAS with BiLSTM-CNN-CRF\textsubscript{AL} (test, cross-domain)}
	\label{fig:system:lcc_atlas_cross_test}
\end{minipage}
\end{figure}

\begin{figure}[!h]\centering
\begin{minipage}{.45\textwidth}\centering
	\includegraphics[width=1\textwidth]{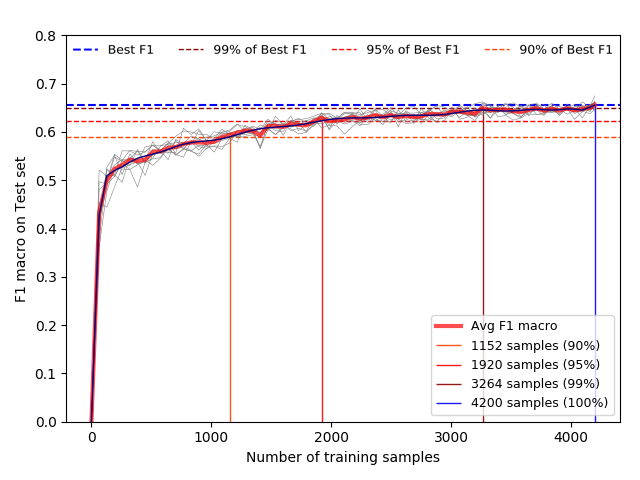}
	\caption{Uncertainty Sampling (Least Confidence) with BiLSTM-CRF\textsubscript{AL} (test, in-domain)}	
	\label{fig:system:lc_cl_lc_in_test}
\end{minipage}\hfill
\begin{minipage}{.45\textwidth}\centering
	\includegraphics[width=1\textwidth]{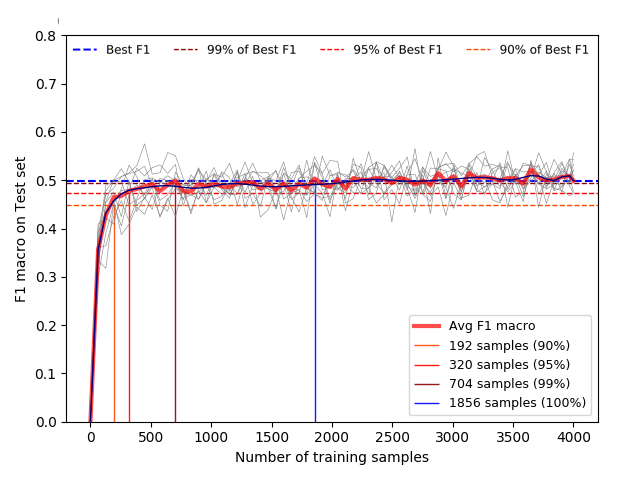}
	\caption{Cluster Uncertainty Sampling (Entropy) with BiLSTM-CRF\textsubscript{AL} (test, cross-domain)}
	\label{fig:system:lc_cl_ent_cross_test}
\end{minipage}
\end{figure}

\begin{figure}[!h]\centering
\begin{minipage}{.45\textwidth}\centering
	\includegraphics[width=1\textwidth]{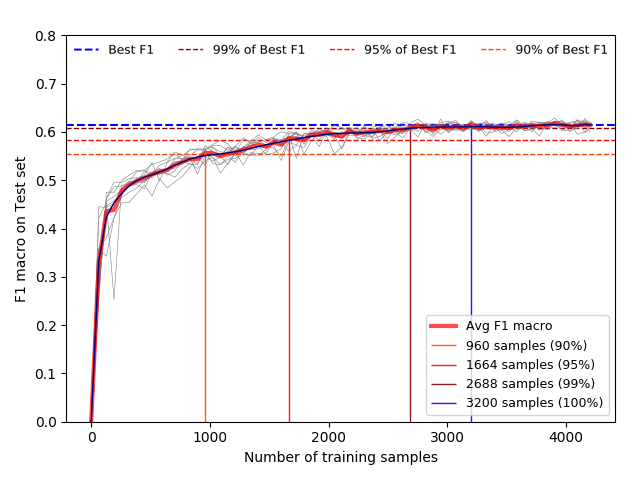}
	\caption[{Uncertainty Sampling (Least Confidence) with LinCRF\textsubscript{AL} (test, in- \protect\linebreak domain)}]{Uncertainty Sampling (Least Confidence) with LinCRF\textsubscript{AL} (test, in-domain)}
	\label{fig:system:lin_lc_in_test}
\end{minipage}\hfill
\begin{minipage}{.45\textwidth}\centering
	\includegraphics[width=1\textwidth]{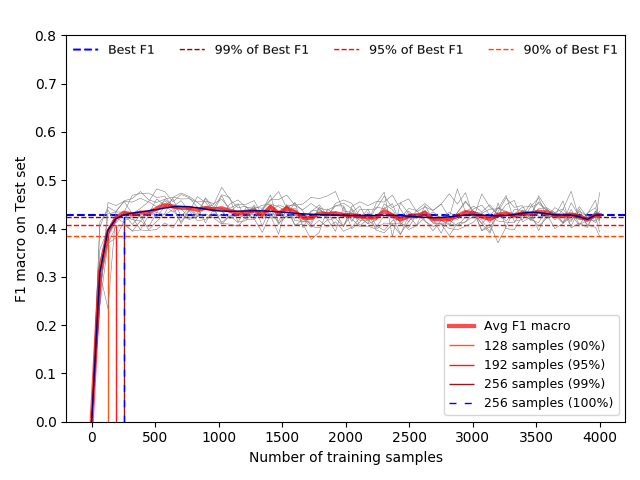}
	\caption{Cluster Representative Sampling with LinCRF\textsubscript{AL} (test, cross-domain)}
	\label{fig:system:lin_cl_rpr_cross_test}
\end{minipage}
\end{figure}

\newpage

\IncMargin{1em}
\begin{algorithm}[H]
\BlankLine
\DontPrintSemicolon

\For{$t = 1, 2,\dots$}
{
\textit{// learn a main model using the current} $\mathcal{L}$\\
$\theta_{M} = \mathbf{train}(\mathcal{L})$\\
\textit{// evaluate main model using the validation set} $\mathcal{V}$\\
\ForEach{$x_{v} \in \mathcal{V}$}
{
$y_{v} = \mathbf{predict}_{\theta_{M}}(x_{v})$\\
}
\textit{// bucket correct and incorrect predictions from} $\mathcal{Y}_{v}$\\
\ForEach{$\langle x_{v}, y_{v} \rangle$}
{
\If{$\mathbf{score}_{\theta_{M}}(y_{v}) = correct$}
{
$\mathcal{V}_{corr} \cup \langle x_{v}, y_{v} \rangle$
}
\Else
{
$\mathcal{V}_{incorr} \cup \langle x_{v}, y_{v} \rangle$
}
}
\textit{// ATLAS step = adaptively choose unlabelled samples}\\
\textit{// using second model until batch is complete}\\
\For{$b$ \KwTo $B$} 
{
\textit{// learn a second binary model}\\
\textit{// using correct and incorrect predictions from} $\mathcal{Y}_{v}$\\
$\theta_{S} = \mathbf{train}(\mathcal{V}_{corr} \cup \mathcal{V}_{incorr})$\\
\textit{// evaluate second binary model using unlabelled data} $\mathcal{U}$\\
\ForEach{$x_{u} \in \mathcal{U}$}
{
$y_{u} = \mathbf{predict}_{\theta_{S}}(x_{u})$\\
}
\textit{// choose randomly b incorrectly classified samples from} $\mathcal{U}$\\
$x_{b}^{*} = \phi_{R}(x_{u_{incorr}})$\\
\textit{// set binary label from incorrect to correct}\\
\textit{// and add instance to validation set for re-training of binary model}\\
\ForEach{$\langle x_{b}^{*}, y_{b}^{*} \rangle$}
{
$y_{b}^{*} = correct$\\
$\mathcal{V}_{corr} = \mathcal{V}_{corr} \cup \langle x_{b}^{*}, y_{b}^{*} \rangle$\\
}
\textit{// move the labelled queries from} $\mathcal{U}$ \textit{to} $\mathcal{L}$\\
$\mathcal{L} = \mathcal{L} \cup \langle  x_{b}^{*}, \mathbf{label}(x_{b}^{*}) \rangle$\\
$\mathcal{U} = \mathcal{U} - x_{b}^{*}$
}
}
\textbf{end}
\caption{\textsc{ATLAS}($\mathcal{L}, \mathcal{U}, \mathcal{V}, \phi_{R}(\cdot), B$)}
\label{fig:related:pool_algo}
\end{algorithm}
\DecMargin{1em}

\newpage 
\noindent \textbf{Data Analysis – In-Domain} \\[0.2em]

\begin{figure}[!h]\centering
\begin{minipage}{.45\textwidth}\centering
	\includegraphics[width=1\textwidth]{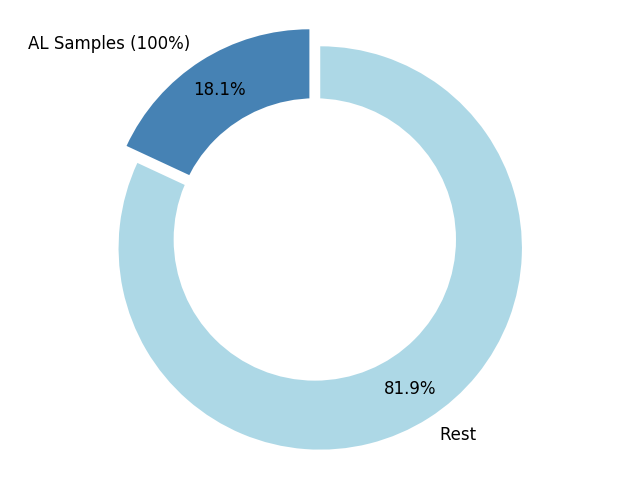}
	\caption{Distribution of AL samples and unlabelled data (in-domain) at $100\%$ threshold for BiLSTM-CNN-CRF\textsubscript{AL} with ATLAS}	
	\label{fig:appendix5:lstm_cnn_crf_inter_rest_inner}
\end{minipage}\hfill
\begin{minipage}{.45\textwidth}\centering
	\includegraphics[width=1\textwidth]{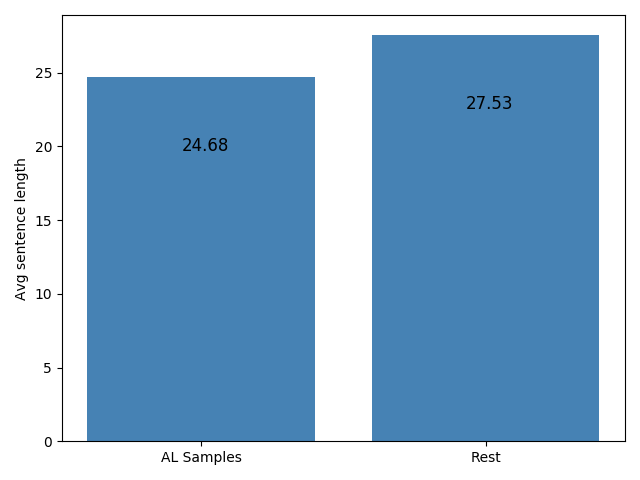}
	\caption{Average sentence length of AL samples and unlabelled data (in-domain) at $100\%$ threshold for BiLSTM-CNN-CRF\textsubscript{AL} with ATLAS}
	\label{fig:appendix5:lstm_cnn_crf_sentlen_inner}
\end{minipage}
\end{figure}

\begin{figure}[!h]\centering
\begin{minipage}{.45\textwidth}\centering
	\includegraphics[width=1\textwidth]{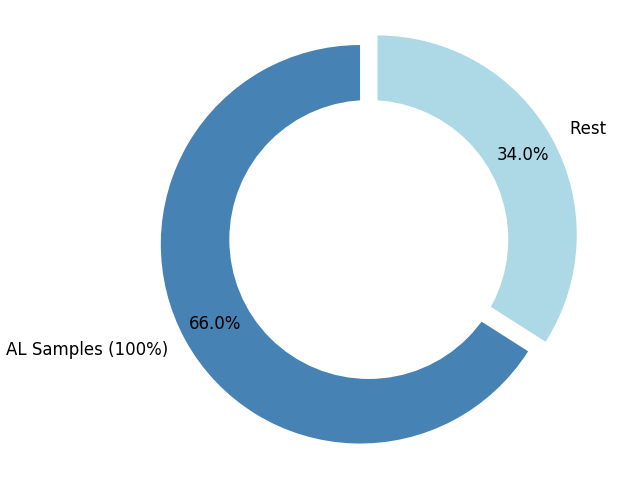}
	\caption[{Distribution of AL samples and unlabelled data (in-domain) at $100\%$ threshold for BiLSTM-CRF\textsubscript{AL} with Uncertainty Sampling (Least \protect\linebreak Confidence)}]{Distribution of AL samples and unlabelled data (in-domain) at $100\%$ threshold for BiLSTM-CRF\textsubscript{AL} with Uncertainty Sampling (Least Confidence)}	
	\label{fig:appendix5:lstm_crf_inter_rest_inner}
\end{minipage}\hfill
\begin{minipage}{.45\textwidth}\centering
	\includegraphics[width=1\textwidth]{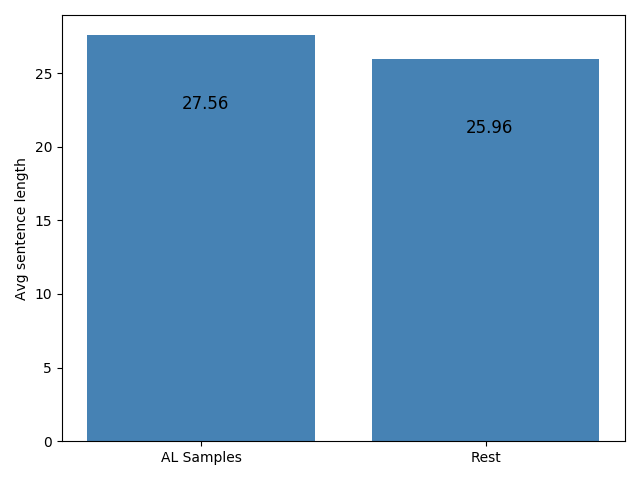}
	\caption{Average sentence length of AL samples and unlabelled data (in-domain) at $100\%$ threshold for BiLSTM-CRF\textsubscript{AL} with Uncertainty Sampling (Least Confidence)}
	\label{fig:appendix5:lstm_crf_sentlen_inner}
\end{minipage}
\end{figure}

\begin{figure}[!h]\centering
\begin{minipage}{.45\textwidth}\centering
	\includegraphics[width=1\textwidth]{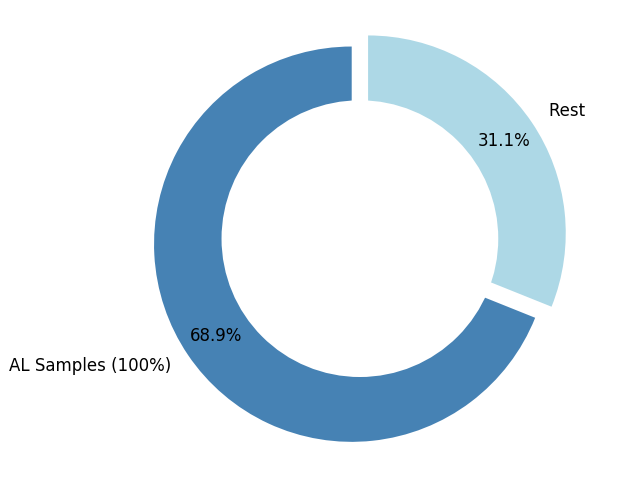}
	\caption{Distribution of AL samples and unlabelled data (in-domain) at $100\%$ threshold for LinCRF\textsubscript{AL} with Uncertainty Sampling (Least Confidence)}	
	\label{fig:appendix5:lincrf_inter_rest_inner}
\end{minipage}\hfill
\begin{minipage}{.45\textwidth}\centering
	\includegraphics[width=1\textwidth]{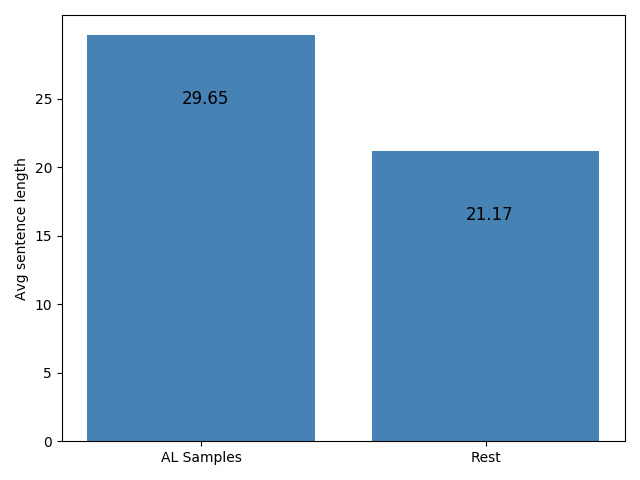}
	\caption{Average sentence length of AL samples and unlabelled data (in-domain) at $100\%$ threshold for LinCRF\textsubscript{AL} with Uncertainty Sampling (Least Confidence)}
	\label{fig:appendix5:lincrf_sentlen_inner}
\end{minipage}
\end{figure}

\begin{figure}[!h]\centering
\begin{minipage}{.45\textwidth}\centering
	\includegraphics[width=1\textwidth]{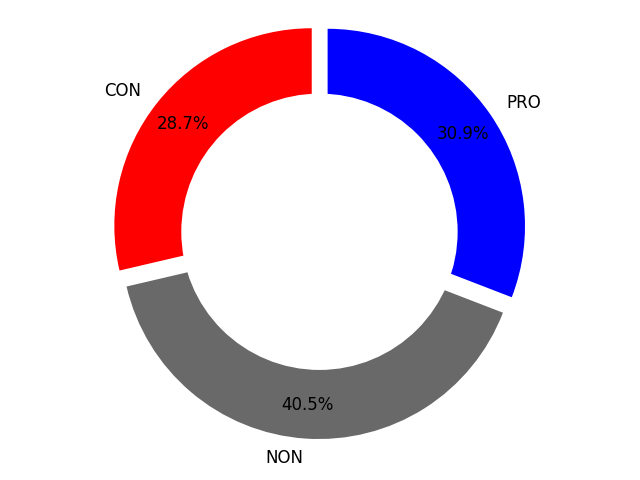}
	\caption{Distribution of labels at token level for AL samples (in-domain) at $100\%$ threshold for BiLSTM-CNN-CRF\textsubscript{AL} with ATLAS}	
	\label{fig:appendix5:lstm_cnn_crf_tokenlabels_inter_inner}
\end{minipage}\hfill
\begin{minipage}{.45\textwidth}\centering
	\includegraphics[width=1\textwidth]{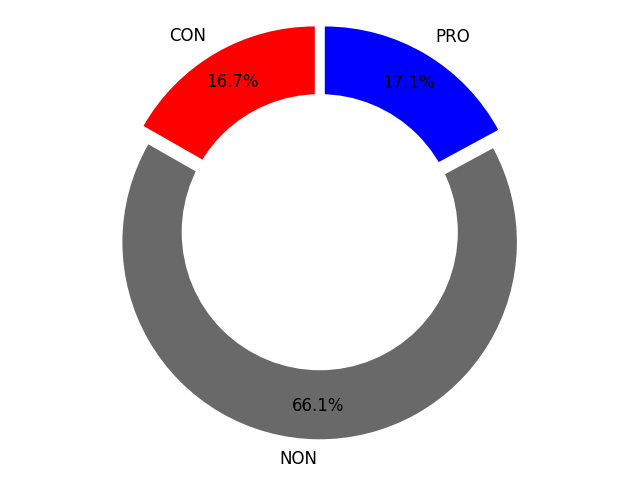}
	\caption{Distribution of labels at token level for unlabelled data (in-domain) at $100\%$ threshold for BiLSTM-CNN-CRF\textsubscript{AL} with ATLAS}
	\label{fig:appendix5:lstm_cnn_crf_tokenlabels_rest_inner}
\end{minipage}
\end{figure}

\begin{figure}[!h]\centering
\begin{minipage}{.45\textwidth}\centering
	\includegraphics[width=1\textwidth]{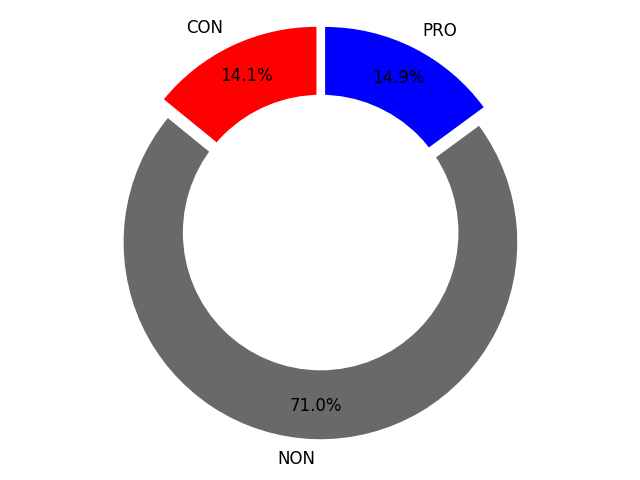}
	\caption[{Distribution of labels at token level for AL samples (in-domain) at $100\%$ threshold for BiLSTM-CRF\textsubscript{AL} with Uncertainty Sampling (Least \protect\linebreak Confidence)}]{Distribution of labels at token level for AL samples (in-domain) at $100\%$ threshold for BiLSTM-CRF\textsubscript{AL} with Uncertainty Sampling (Least Confidence)}	
	\label{fig:appendix5:lstm_crf_tokenlabels_inter_inner}
\end{minipage}\hfill
\begin{minipage}{.45\textwidth}\centering
	\includegraphics[width=1\textwidth]{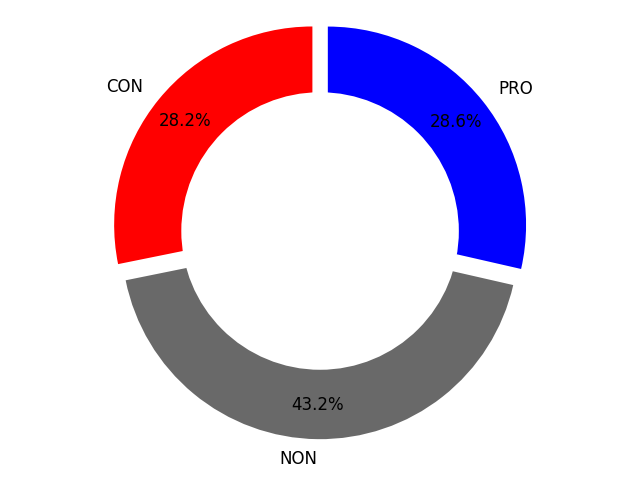}
	\caption{Distribution of labels at token level for unlabelled data (in-domain) at $100\%$ threshold for BiLSTM-CRF\textsubscript{AL} with Uncertainty Sampling (Least Confidence)}
	\label{fig:appendix5:lstm_crf_tokenlabels_rest_inner}
\end{minipage}
\end{figure}

\begin{figure}[!h]\centering
\begin{minipage}{.45\textwidth}\centering
	\includegraphics[width=1\textwidth]{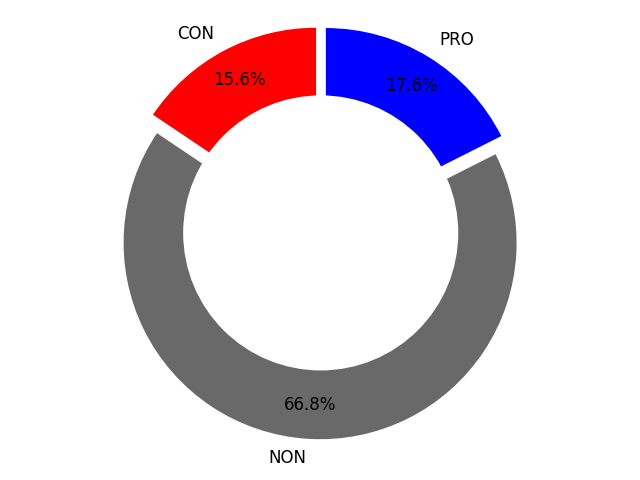}
	\caption{Distribution of labels at token level for AL samples (in-domain) at $100\%$ threshold for LinCRF\textsubscript{AL} with Uncertainty Sampling (Least Confidence)}	
	\label{fig:appendix5:lincrf_tokenlabels_inter_inner}
\end{minipage}\hfill
\begin{minipage}{.45\textwidth}\centering
	\includegraphics[width=1\textwidth]{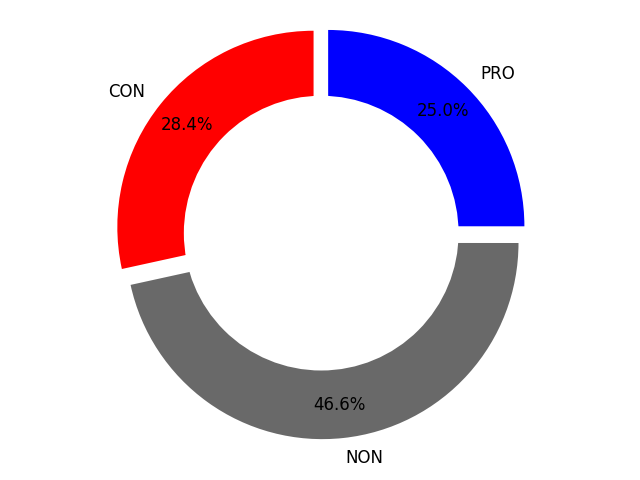}
	\caption{Distribution of labels at token level for unlabelled data (in-domain) at $100\%$ threshold for LinCRF\textsubscript{AL} with Uncertainty Sampling (Least Confidence)}
	\label{fig:appendix5:lincrf_tokenlabels_rest_inner}
\end{minipage}
\end{figure}

\begin{figure}[!h]\centering
\begin{minipage}{.45\textwidth}\centering
	\includegraphics[width=1\textwidth]{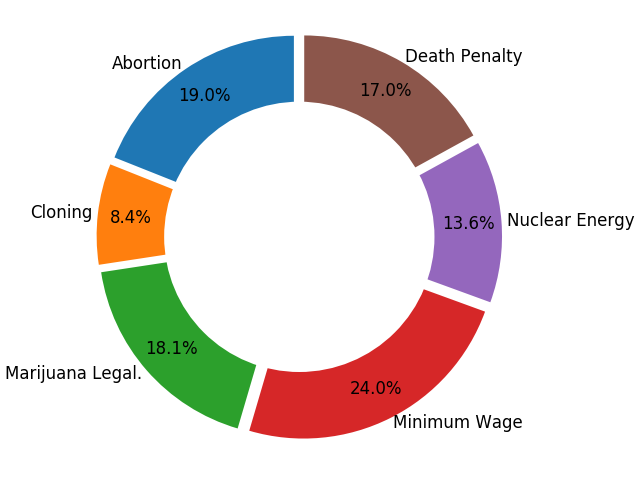}
	\caption{Distribution of topics for AL samples (in-domain) at $100\%$ threshold for BiLSTM-CNN-CRF\textsubscript{AL} with ATLAS}	
	\label{fig:appendix5:lstm_cnn_crf_topics_inter_inner}
\end{minipage}\hfill
\begin{minipage}{.45\textwidth}\centering
	\includegraphics[width=1\textwidth]{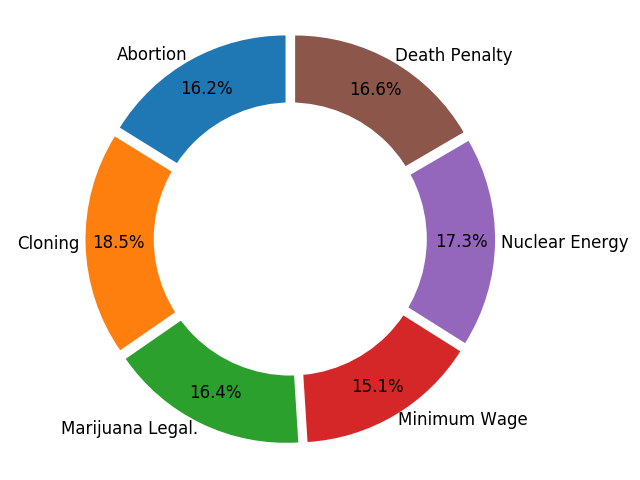}
	\caption{Distribution of topics for unlabelled data (in-domain) at $100\%$ threshold for BiLSTM-CNN-CRF\textsubscript{AL} with ATLAS}
	\label{fig:appendix5:lstm_cnn_crf_topics_rest_inner}
\end{minipage}
\end{figure}

\begin{figure}[!h]\centering
\begin{minipage}{.45\textwidth}\centering
	\includegraphics[width=1\textwidth]{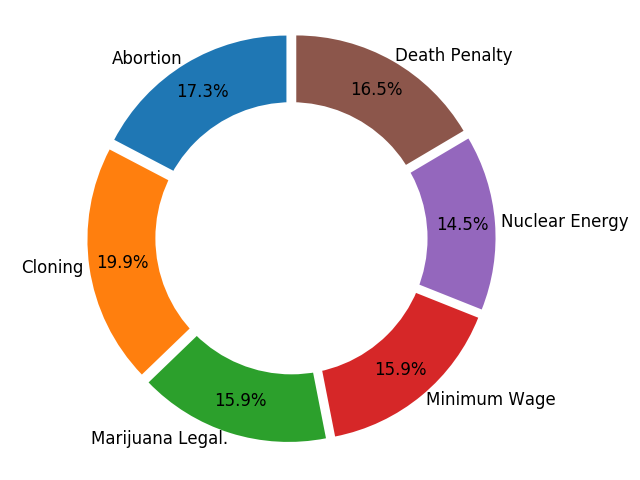}
	\caption{Distribution of topics for AL samples (in-domain) at $100\%$ threshold for BiLSTM-CRF\textsubscript{AL} with Uncertainty Sampling (Least Confidence)}	
	\label{fig:appendix5:lstm_crf_topics_inter_inner}
\end{minipage}\hfill
\begin{minipage}{.45\textwidth}\centering
	\includegraphics[width=1\textwidth]{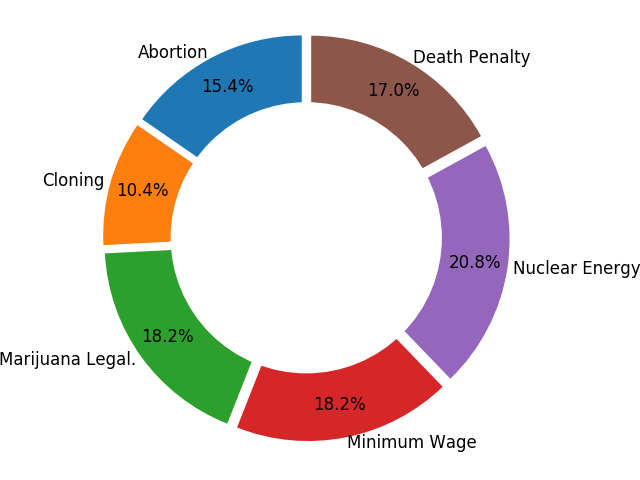}
	\caption{Distribution of topics for unlabelled data (in-domain) at $100\%$ threshold for BiLSTM-CRF\textsubscript{AL} with Uncertainty Sampling (Least Confidence)}
	\label{fig:appendix5:lstm_crf_topics_rest_inner}
\end{minipage}
\end{figure}

\begin{figure}[!h]\centering
\begin{minipage}{.45\textwidth}\centering
	\includegraphics[width=1\textwidth]{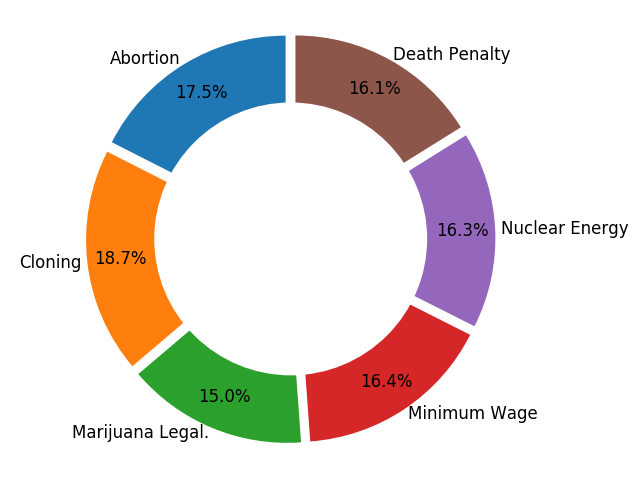}
	\caption{Distribution of topics for AL samples (in-domain) at $100\%$ threshold for LinCRF\textsubscript{AL} with Uncertainty Sampling (Least Confidence)}	
	\label{fig:appendix5:lincrf_topics_inter_inner}
\end{minipage}\hfill
\begin{minipage}{.45\textwidth}\centering
	\includegraphics[width=1\textwidth]{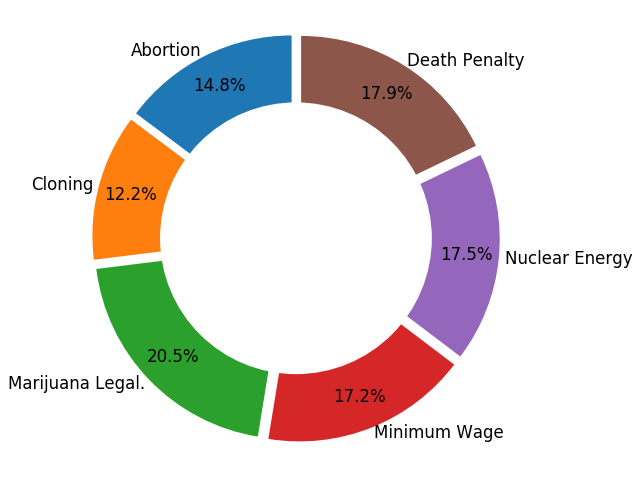}
	\caption{Distribution of topics for unlabelled data (in-domain) at $100\%$ threshold for LinCRF\textsubscript{AL} with Uncertainty Sampling (Least Confidence)}
	\label{fig:appendix5:lincrf_topics_rest_inner}
\end{minipage}
\end{figure}

\cleardoublepage

\noindent \textbf{Data Analysis – Cross-Domain} \\[0.2em]

\begin{figure}[!h]\centering
\begin{minipage}{.45\textwidth}\centering
	\includegraphics[width=1\textwidth]{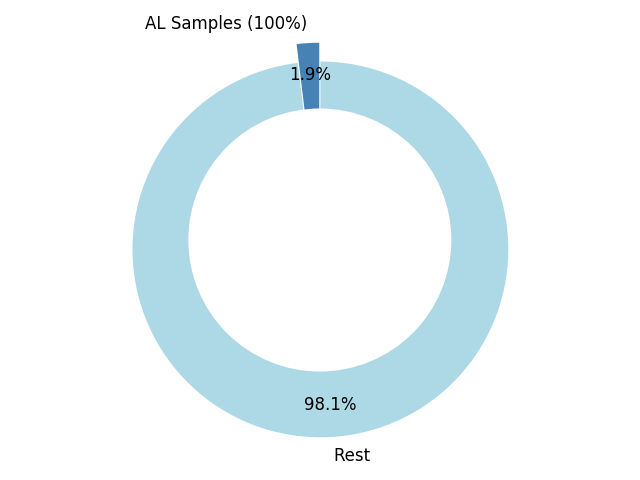}
	\caption{Distribution of AL samples and unlabelled data (cross-domain) at $100\%$ threshold for BiLSTM-CNN-CRF\textsubscript{AL} with ATLAS}	
	\label{fig:appendix5:lstm_cnn_crf_inter_rest_cross}
\end{minipage}\hfill
\begin{minipage}{.45\textwidth}\centering
	\includegraphics[width=1\textwidth]{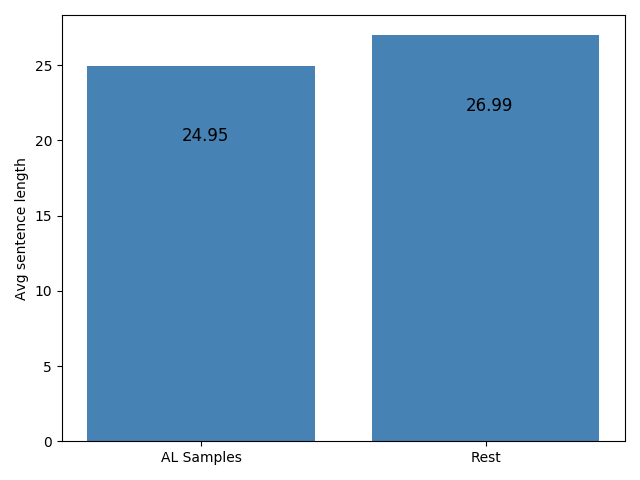}
	\caption{Average sentence length of AL samples and unlabelled data (cross-domain) at $100\%$ threshold for BiLSTM-CNN-CRF\textsubscript{AL} with ATLAS}
	\label{fig:appendix5:lstm_cnn_crf_sentlen_cross}
\end{minipage}
\end{figure}

\begin{figure}[!h]\centering
\begin{minipage}{.45\textwidth}\centering
	\includegraphics[width=1\textwidth]{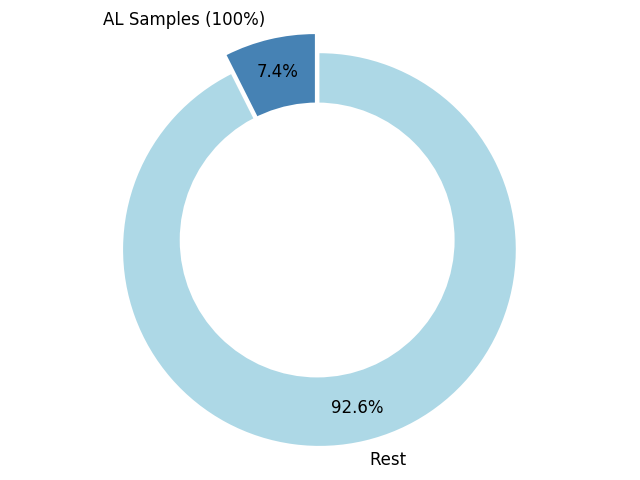}
	\caption[{Distribution of AL samples and unlabelled data (cross-domain) at $100\%$ threshold for BiLSTM-CRF\textsubscript{AL} with Cluster Uncertainty Sampling \protect\linebreak (Entropy)}]{Distribution of AL samples and unlabelled data (cross-domain) at $100\%$ threshold for BiLSTM-CRF\textsubscript{AL} with Cluster Uncertainty Sampling (Entropy)}	
	\label{fig:appendix5:lstm_crf_inter_rest_cross}
\end{minipage}\hfill
\begin{minipage}{.45\textwidth}\centering
	\includegraphics[width=1\textwidth]{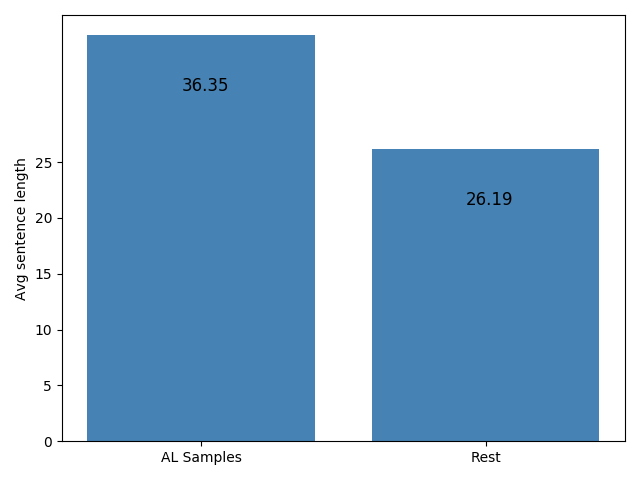}
	\caption{Average sentence length of AL samples and unlabelled data (cross-domain) at $100\%$ threshold for BiLSTM-CRF\textsubscript{AL} with Cluster Uncertainty Sampling (Entropy)}
	\label{fig:appendix5:lstm_crf_sentlen_cross}
\end{minipage}
\end{figure}

\begin{figure}[!h]\centering
\begin{minipage}{.45\textwidth}\centering
	\includegraphics[width=1\textwidth]{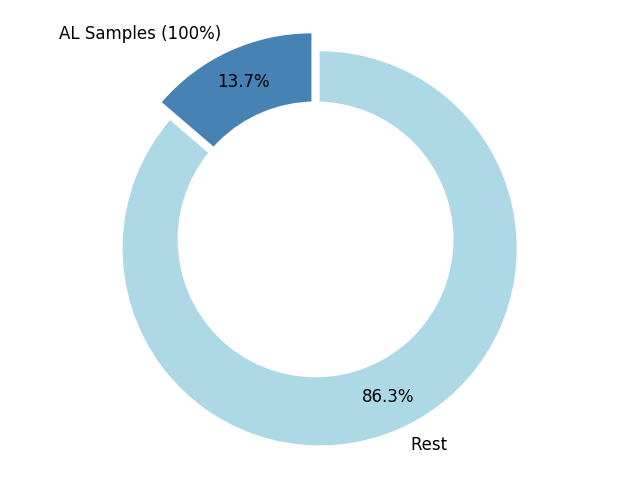}
	\caption{Distribution of AL samples and unlabelled data (cross-domain) at $100\%$ threshold for LinCRF\textsubscript{AL} with Cluster Representative Sampling}	
	\label{fig:appendix5:lincrf_inter_rest_cross}
\end{minipage}\hfill
\begin{minipage}{.45\textwidth}\centering
	\includegraphics[width=1\textwidth]{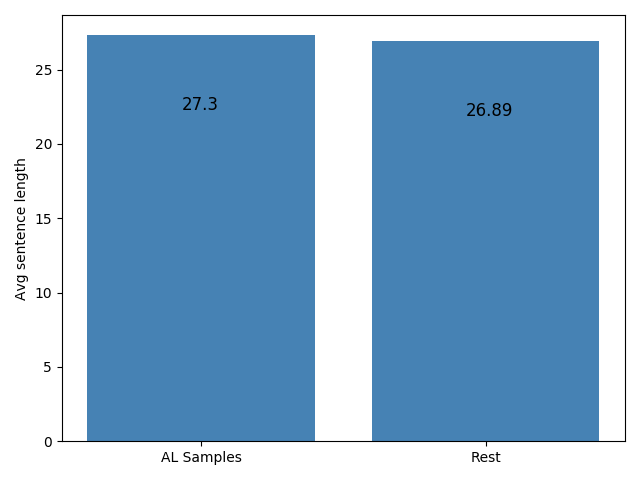}
	\caption{Average sentence length of AL samples and unlabelled data (cross-domain) at $100\%$ threshold for LinCRF\textsubscript{AL} with Cluster Representative Sampling}
	\label{fig:appendix5:lincrf_sentlen_cross}
\end{minipage}
\end{figure}

\begin{figure}[!h]\centering
\begin{minipage}{.45\textwidth}\centering
	\includegraphics[width=1\textwidth]{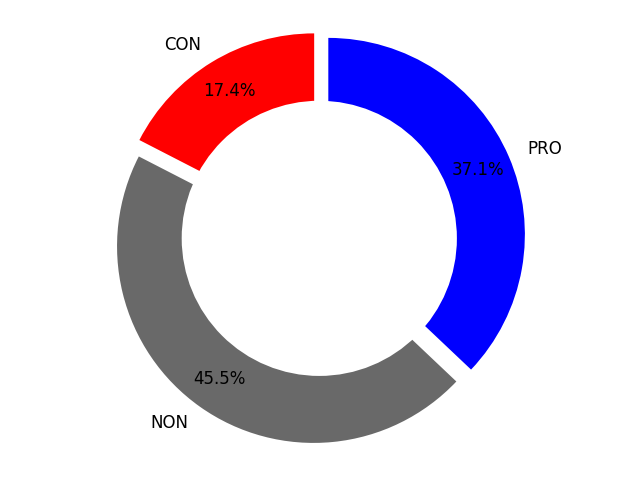}
	\caption{Distribution of labels at token level for AL samples (cross-domain) at $100\%$ threshold for BiLSTM-CNN-CRF\textsubscript{AL} with ATLAS}	
	\label{fig:appendix5:lstm_cnn_crf_tokenlabels_inter_cross}
\end{minipage}\hfill
\begin{minipage}{.45\textwidth}\centering
	\includegraphics[width=1\textwidth]{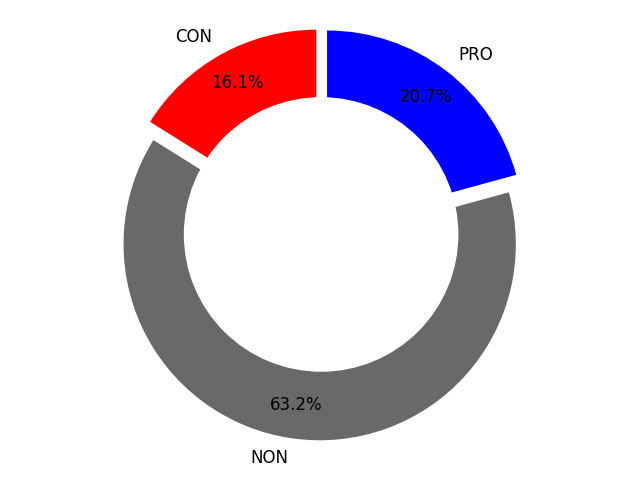}
	\caption{Distribution of labels at token level for unlabelled data (cross-domain) at $100\%$ threshold for BiLSTM-CNN-CRF\textsubscript{AL} with ATLAS}
	\label{fig:appendix5:lstm_cnn_crf_tokenlabels_rest_cross}
\end{minipage}
\end{figure}

\begin{figure}[!h]\centering
\begin{minipage}{.45\textwidth}\centering
	\includegraphics[width=1\textwidth]{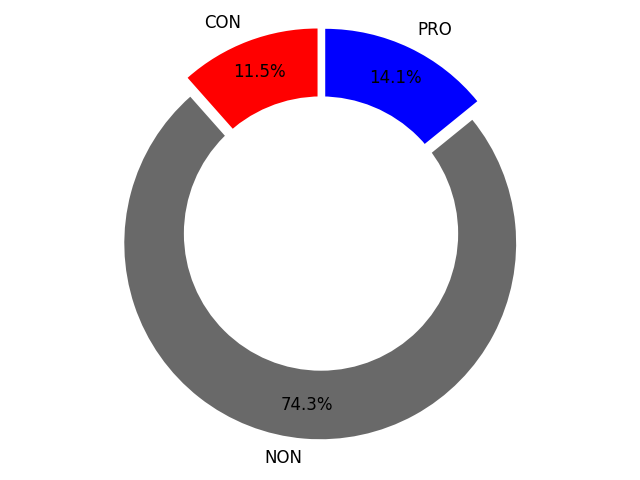}
	\caption{Distribution of labels at token level for AL samples (cross-domain) at $100\%$ threshold for BiLSTM-CRF\textsubscript{AL} with Cluster Uncertainty Sampling (Entropy)}	
	\label{fig:appendix5:lstm_crf_tokenlabels_inter_cross}
\end{minipage}\hfill
\begin{minipage}{.45\textwidth}\centering
	\includegraphics[width=1\textwidth]{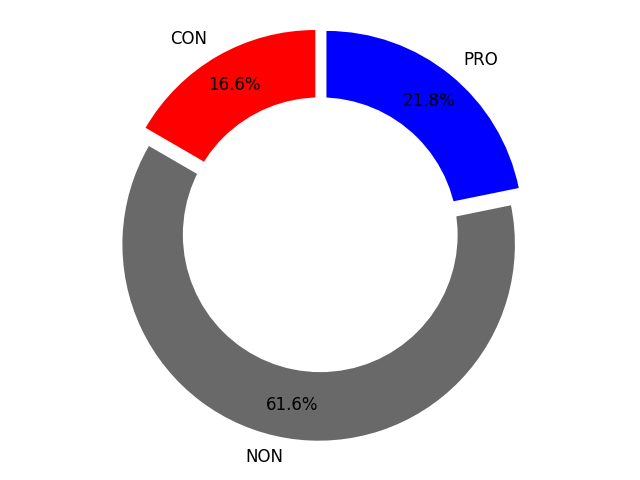}
	\caption{Distribution of labels at token level for unlabelled data (cross-domain) at $100\%$ threshold for BiLSTM-CRF\textsubscript{AL} with Cluster Uncertainty Sampling (Entropy)}
	\label{fig:appendix5:lstm_crf_tokenlabels_rest_cross}
\end{minipage}
\end{figure}

\begin{figure}[!h]\centering
\begin{minipage}{.45\textwidth}\centering
	\includegraphics[width=1\textwidth]{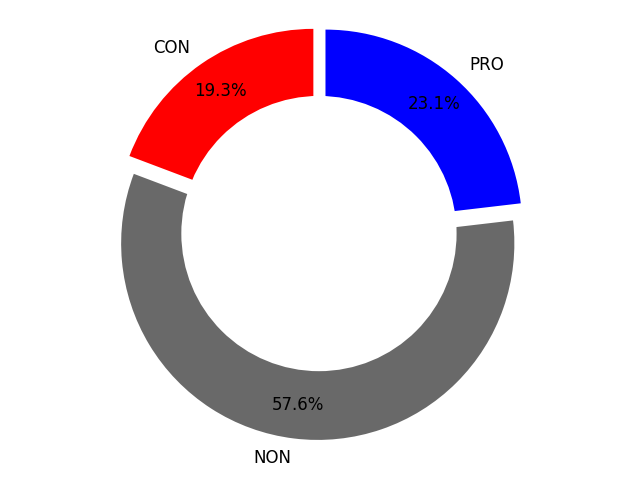}
	\caption{Distribution of labels at token level for AL samples (cross-domain) at $100\%$ threshold for LinCRF\textsubscript{AL} with Cluster Representative Sampling}	
	\label{fig:appendix5:lincrf_tokenlabels_inter_cross}
\end{minipage}\hfill
\begin{minipage}{.45\textwidth}\centering
	\includegraphics[width=1\textwidth]{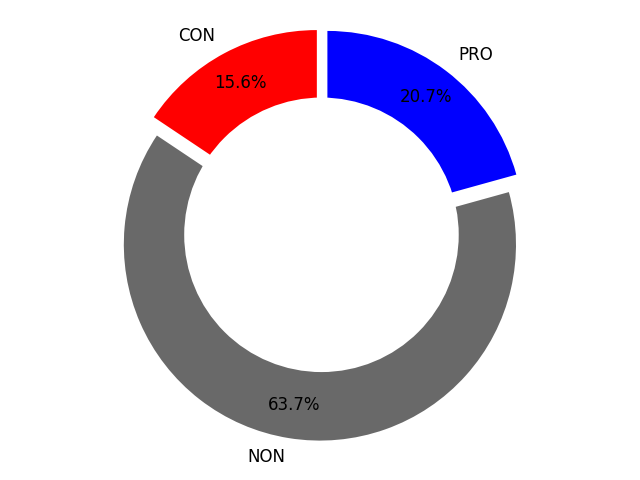}
	\caption{Distribution of labels at token level for unlabelled data (cross-domain) at $100\%$ threshold for LinCRF\textsubscript{AL} with Cluster Representative Sampling}
	\label{fig:appendix5:lincrf_tokenlabels_rest_cross}
\end{minipage}
\end{figure}

\begin{figure}[!h]\centering
\begin{minipage}{.45\textwidth}\centering
	\includegraphics[width=1\textwidth]{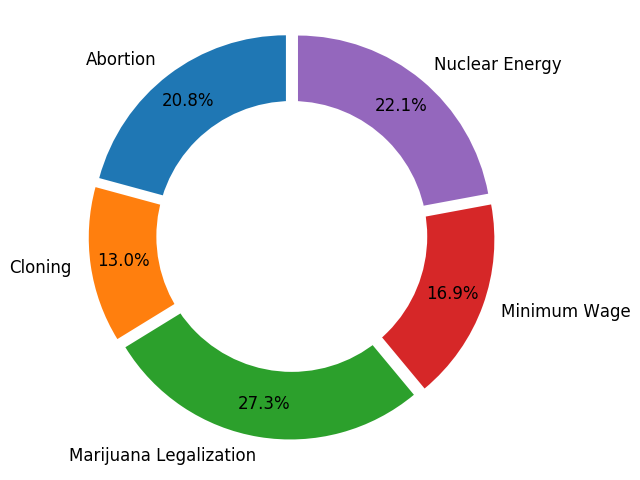}
	\caption{Distribution of topics for AL samples (cross-domain) at $100\%$ threshold for BiLSTM-CNN-CRF\textsubscript{AL} with ATLAS}	
	\label{fig:appendix5:lstm_cnn_crf_topics_inter_cross}
\end{minipage}\hfill
\begin{minipage}{.45\textwidth}\centering
	\includegraphics[width=1\textwidth]{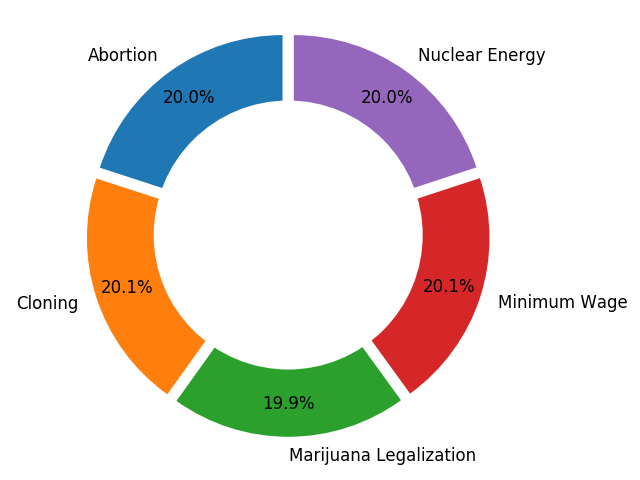}
	\caption{Distribution of topics for unlabelled data (cross-domain) at $100\%$ threshold for BiLSTM-CNN-CRF\textsubscript{AL} with ATLAS}
	\label{fig:appendix5:lstm_cnn_crf_topics_rest_cross}
\end{minipage}
\end{figure}

\begin{figure}[!h]\centering
\begin{minipage}{.45\textwidth}\centering
	\includegraphics[width=1\textwidth]{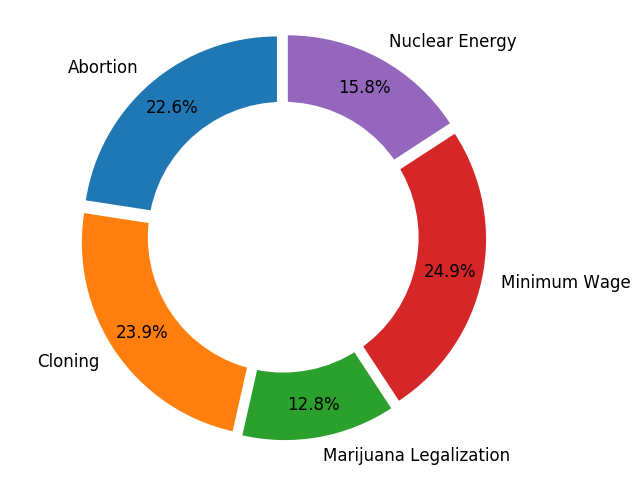}
	\caption{Distribution of topics for AL samples (cross-domain) at $100\%$ threshold for BiLSTM-CRF\textsubscript{AL} with Cluster Uncertainty Sampling (Entropy)}	
	\label{fig:appendix5:lstm_crf_topics_inter_cross}
\end{minipage}\hfill
\begin{minipage}{.45\textwidth}\centering
	\includegraphics[width=1\textwidth]{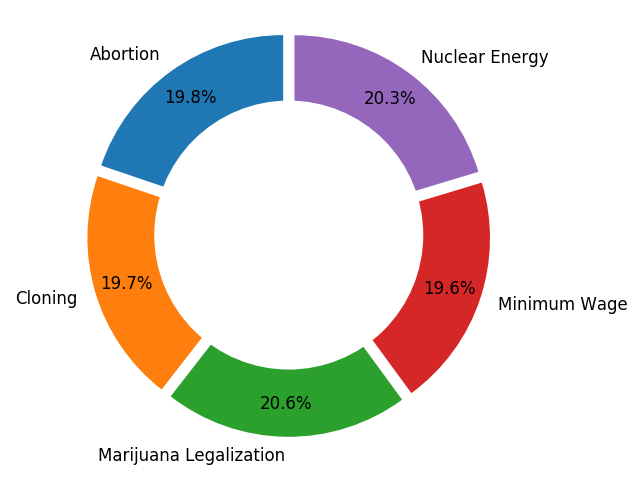}
	\caption{Distribution of topics for unlabelled data (cross-domain) at $100\%$ threshold for BiLSTM-CRF\textsubscript{AL} with Cluster Uncertainty Sampling (Entropy)}
	\label{fig:appendix5:lstm_crf_topics_rest_cross}
\end{minipage}
\end{figure}

\begin{figure}[!h]\centering
\begin{minipage}{.45\textwidth}\centering
	\includegraphics[width=1\textwidth]{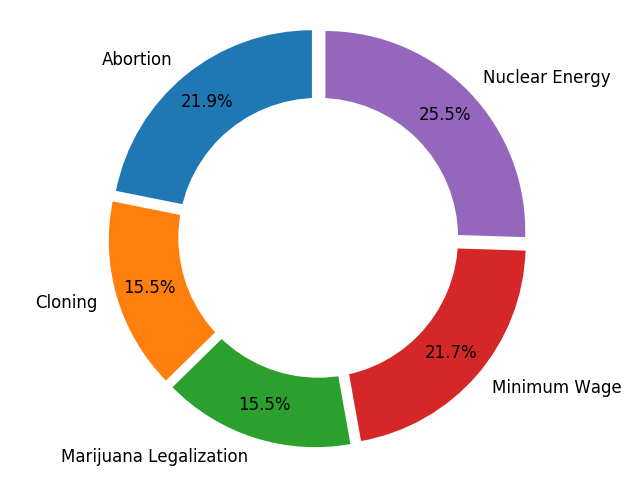}
	\caption{Distribution of topics for AL samples (cross-domain) at $100\%$ threshold for LinCRF\textsubscript{AL} with Cluster Representative Sampling}	
	\label{fig:appendix5:lincrf_topics_inter_cross}
\end{minipage}\hfill
\begin{minipage}{.45\textwidth}\centering
	\includegraphics[width=1\textwidth]{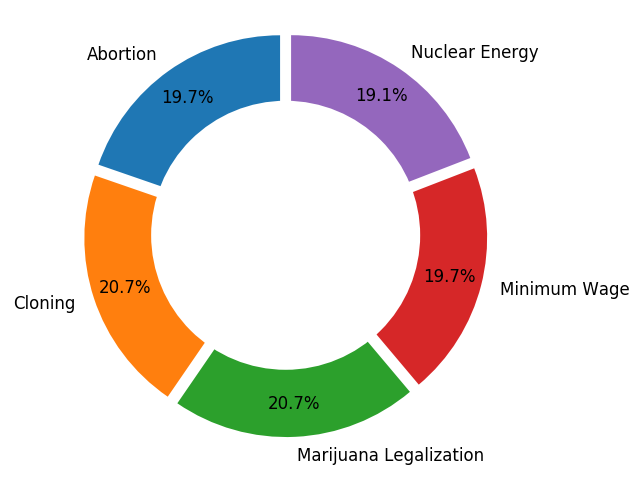}
	\caption{Distribution of topics for unlabelled data (cross-domain) at $100\%$ threshold for LinCRF\textsubscript{AL} with Cluster Representative Sampling}
	\label{fig:appendix5:lincrf_topics_rest_cross}
\end{minipage}
\end{figure}

\end{document}